\documentclass[letterpaper]{article}
\usepackage{UAI_style}
\usepackage[margin=1in]{geometry}

\usepackage{times}
\usepackage[utf8]{inputenc} % allow utf-8 input
\usepackage[T1]{fontenc}    % use 8-bit T1 fonts

\usepackage{url}            % simple URL typesetting
\usepackage{booktabs}       % professional-quality tables
\usepackage{amsfonts}       % blackboard math symbols
\usepackage{nicefrac}       % compact symbols for 1/2, etc.
\usepackage{microtype}      % microtypography
\usepackage{amsthm}
\usepackage{amsmath}
\usepackage{amssymb}
\usepackage{color}
\usepackage{graphicx}
\usepackage{caption}
\usepackage{paralist}
\usepackage{subcaption}
\usepackage{tikz}
\usepackage{cuted}
\usepackage{smallref}
\usepackage{bm}
\usetikzlibrary{arrows}
\usetikzlibrary{positioning,automata,calc,patterns,decorations.pathmorphing,decorations.markings}

\usepackage{todonotes}
\let\todoorig\todo
\renewcommand{\todo}[1]{\todoorig[inline]{#1}}

\renewcommand{\pmb}[1]{\bm{#1}} % use proper bold fonts rather than "poor man's bold"

\newcommand{\intset}{\mathtt{Dyn}}
\newcommand{\dynset}{\mathtt{Dyn}}
\newcommand{\pathset}{\mathtt{Traj}}
\newcommand{\pa}{\mathtt{pa}}
\newcommand{\doop}{\mathtt{do}}
\newcommand{\Eta}{\pmb{\eta}}
\newcommand{\Zeta}{\pmb{\zeta}}

\makeatletter
\newcommand\footnoteref[1]{\protected@xdef\@thefnmark{\ref{#1}}\@footnotemark}
\makeatother

\newtheorem{definition}{Definition}

\newtheorem{theorem}{Theorem}

\newtheorem{corollary}{Corollary}
\newtheorem{example}{Example}
\newenvironment{contexample}{
   \addtocounter{example}{-1} \begin{example}[continued]}{
   \end{example}}
\theoremstyle{remark}
\newtheorem{remark}{Remark}
\usepackage{natbib}
\setcitestyle{authoryear}

\newcommand\eref[1]{(\ref{#1})}

\title{From Deterministic ODEs to Dynamic Structural Causal Models}

\author{ {\bf Paul K.~Rubenstein\thanks{\;Also affiliated with Max Planck Institute for Intelligent Systems, T\"ubingen.}} \\
Department of Engineering \\
University of Cambridge\\
United Kingdom\\
\texttt{pkr23@cam.ac.uk} \\
\And
{\bf Stephan Bongers}  \\
Informatics Institute \\
University of Amsterdam \\
The Netherlands\\
\texttt{S.R.Bongers@uva.nl}\\
\And
{\bf Bernhard Sch\"olkopf}   \\
Max-Planck Institute for\\
Intelligent Systems, T\"ubingen    \\
Germany\\
\texttt{bs@tue.mpg.de}\\
\And
{\bf Joris M.\ Mooij}   \\
Informatics Institute \\
University of Amsterdam \\
The Netherlands\\
\texttt{J.M.Mooij@uva.nl}\\
}

\begin{document}

\maketitle
 
\begin{abstract}\normalsize
Structural Causal Models are widely used in causal modelling, but how they relate to other modelling tools is poorly understood.
In this paper we provide a novel perspective on the relationship between Ordinary Differential Equations and Structural Causal Models.
We show how, under certain conditions, the asymptotic behaviour of an Ordinary Differential Equation under non-constant interventions can be modelled using Dynamic Structural Causal Models.
In contrast to earlier work, we study not only the effect of interventions on equilibrium states; rather, we model asymptotic behaviour that is \emph{dynamic} under interventions that vary in time, and include as a special case the study of static equilibria. 
\end{abstract}

\vspace{-0.15cm}
\section{INTRODUCTION}

%ODEs describe the physics of a system, and modelling causality here is `straightforward', though can result in solutions that are arbitrarily complex. Statistics says nothing about causality. SCMs are somewhere in between; subject to certain restrictions on what we are trying to model, SCMs give a way to do causal modelling in a `simpler space' than at the level of ODEs. [Mooij] shows how if you make the assumption of equilibrium and constant intervention, you can derive an SCM to model the effect on the equilibrium. In this paper, we relax this constraint and show a more general condition on the ODE and (non-constant) interventions we want to model that, if satisfied, allows us to derive SCMs to causally model the interventions.

Ordinary Differential Equations (ODEs) provide a universal language to describe deterministic systems via equations that determine how variables change in time as a function of other variables. They provide an immensely popular and highly successful modelling framework, with applications in many diverse disciplines, such as physics, chemistry, biology, and economy. They are \emph{causal} in the sense that at least in principle they allow us to reason about interventions: any external intervention in a system---e.g., moving an object by applying a force---can be modelled using modified differential equations by, for instance, including suitable forcing terms. In practice, of course, this may be arbitrarily difficult.

%Statistical dependences in the world are generally the outcome of time evolution governed by differential equations. For instance, to produce a dataset of labelled handwritten digits, we might instruct a writer to produce images of digits belonging to specified classes. This instantiates a complex dynamical process involving neural networks in the brain as well as muscles. At the end, we are left with statistical dependences that machine learning methods can exploit. However, the causal structure that was present in the differential equations is lost. 

Structural Causal Models (SCMs, also known as Structural Equation Models) are another language capable of describing causal relations and interventions and have been widely applied in the social sciences, economics, genetics and neuroscience \citep{Pearl2009, bollen2014structural}. One of the successes of SCMs over other causal frameworks such as causal Bayesian networks, for instance, has been their ability to express cyclic causal models \citep{spirtes1995directed,mooij2011causal,hyttinen2012learning,voortman2012learning,lacerda2012discovering,Bongers++_1611.06221v2}.
%druzdzel1993causality,

We view SCMs as an intermediate level of description between the highly expressive differential equation models and the probabilistic, non-causal models typically used in machine learning and statistics. This intermediate level of description ideally retains the benefits of a data-driven statistical approach while still allowing a limited set of causal statements about the effect of interventions. While it is well understood how an SCM induces a statistical model \citep{Bongers++_1611.06221v2}, much less is known about how a differential equation model---our most fundamental level of modelling---can imply an SCM in the first place.
%A particular advantage of them over other causal models (such as Bayesian Networks,...) is that they naturally accommodate cases of cyclic causal dependency, ie when feedback loops exist \citep{hyttinen2012learning,  mooij2011causal, voortman2012learning, lacerda2012discovering,spirtes1995directed}.
%Building on Iwasaki and Simon \citep{iwasaki1994causality},\Bernhard{did we really build on their work -- our citation of Iwasaki and Simon in the UAI paper was more casual} 
This is an important question because if we are to have models of a system on different levels of complexity, we should understand how they relate and the conditions under which they are consistent with one another. 
%While this relationship may not be practically feasible to study in a mathematical framework in the case of connected neurons giving rise to handwritten digits, we can study this in the case of simpler dynamical systems.

Indeed, recent work has begun to address the question of how SCMs arise naturally from more fundamental models by showing how, under strong assumptions, SCMs can be derived from an underlying discrete time difference equation or continuous time ODE \citep{iwasaki1994causality,dash2005restructuring,lacerda2012discovering,voortman2012learning,MooJanSch13,SokolHansen2014}. With the exception of \citep{voortman2012learning} and \citep{SokolHansen2014}, each of these methods assume that the dynamical system comes to a static equilibrium that is independent of initial conditions, with the derived SCM describing how this equilibrium changes under intervention. More recently, the more general case in which the equilibrium state may depend on the initial conditions has been addressed \citep{BongersMooij_1803.08784,BlomMooij_1805.06539}.

If the assumption that the system reaches a static equilibrium is reasonable for a particular system under study, the SCM framework can be useful. Although the derived SCM then lacks information about the (possibly rich) transient dynamics of the system, if the system equilibrates quickly then the description of the system as an SCM may be a more convenient and compact representation of the causal structure of interest. By making assumptions on the dynamical system and the interventions being made, the SCM effectively allows us to reason about a `higher level' qualitative description of the dynamics---in this case, the equilibrium states.

There are, however, two major limitations that stem from the equilibrium assumption. First, for many dynamical systems the assumption that the system settles to a unique equilibrium, either in its observational state or under intervention, may be a bad approximation of the actual system dynamics. Second, this framework is only capable of modelling interventions in which a subset of variables are clamped to fixed values (\emph{constant} interventions). Even for rather simple physical systems such as a forced damped simple harmonic oscillator, these assumptions are violated. 

Motivated by these observations, the work presented in this paper tries to answer the following questions:
(i) Can the SCM framework be extended to model systems that do not converge to an equilibrium? (ii)
If so, what assumptions need to be made on the ODE and interventions so that this is possible? 
Since SCMs are used in a variety of situations in which the equilibrium assumption does not necessarily hold, we view these questions as important in order to understand when they are indeed theoretically grounded as modelling tools.
%While we do not provide a complete characterisation of when SCMs can be used to describe non-equilibrating systems,\footnote{Indeed, this is an ill-posed question that could be made mathematically precise in many ways.} 
The main contribution of this paper is to show that the answer to the first question is `Yes' and to provide sufficient conditions for the second. We do this by extending the SCM framework to encompass time-dependent dynamics and interventions and studying how such objects can arise from ODEs. We refer to this as a \emph{Dynamic SCM (DSCM)} to distinguish it from the static equilibrium case for the purpose of exposition, but note that this is conceptually the same as an SCM on a fundamental level.
Our construction draws inspiration from the approach of \cite{MooJanSch13}, that was recently generalized to also
incorporate the stochastic setting \citep{BongersMooij_1803.08784}. Here, we adapt the approach by replacing the static equilibrium states by
continuous-time \emph{trajectories}, considering two trajectories as equivalent if they do not differ asymptotically.

Note that whilst this paper applies a causal perspective to the study of dynamical systems, the goal of this paper is not to derive a learning algorithm which can be applied to time series data. In this sense, we view our main results as `orthogonal' to methods such as Granger causality \citep{granger1969investigating} and difference-in-differences \citep{card1993minimum} which aim to infer causal effects given time-series observations of a system. 
We envision that DSCMs may be used for causal analysis of dynamical systems that undergo periodic motion. Although these systems have been mostly ignored so far in the field of causal discovery, they have been studied extensively in the field of control theory. 
%One example of an interesting finding is that certain chemical reactions may proceed more efficiently when the concentrations of the reactants are driven to undergo certain oscillations than when they are controlled at steady levels.
Some examples of systems that naturally exhibit oscillatory stationary states and where our framework may be applicable are EEG signals, circadian signals, seasonal influences, chemical oscillations, electric circuits, aerospace vehicles, and satellite control. We refer the reader to \citep{bittanti2009periodic} for more details on these application areas from the perspective of periodic control theory. 

Since the DSCM derived for a simple harmonic oscillator (see Example \ref{example:dscm}) is already quite complex, we leave the task of deriving methods that estimate the parameters from
data for future work.
Rather, our current work presents a first necessary theoretical step that needs to be done
before applications of this theory can be developed, enabling the development of data-driven causal discovery and prediction methods for oscillatory systems, and possibly even more general systems, down the road. 

The remainder of this paper is organised as follows. In Section~\ref{section:ode}, we introduce notation to describe ODEs. In Section~\ref{section:interventions}, we describe how to apply the notion of an intervention on an ODE to the dynamic case. In Section~\ref{section:dynamic-stability}, we define regularity conditions on the asymptotic behaviour of an ODE under a set of interventions. In Section~\ref{section:dscm}, we present our main result: subject to conditions on the dynamical system and interventions being modelled, a \emph{Dynamic SCM} can be derived that allows one to reason about how the asymptotic dynamics change under interventions on variables in the system. We conclude in Section~\ref{section:discussion}.

\vspace{-0.15cm}
\section{ORDINARY DIFFERENTIAL EQUATIONS}\label{section:ode}
\vspace{-0.1cm}
%\begin{itemize}
%\item Give description of ODE systems. In particular, the fact that we don't separate out variables and their derivatives, but rather have a single equation per variable, since we want to have possibly complex interventions.
%\end{itemize}

Let ${\mathcal{I}= \{1,\ldots,D\}}$ be a set of variable labels. Consider time-indexed variables ${X_i(t) \in \mathcal{R}_i}$ for ${i \in \mathcal{I}}$,  where ${\mathcal{R}_i \subseteq \mathbb{R}}$ and ${t\in\mathbb{R}_{\geq 0} = [0,\infty)}$. For ${I \subseteq \mathcal{I}}$, we write ${\mathbf{X}_I(t) \in \prod_{i\in I} \mathcal{R}_i}$ for the tuple of variables ${(X_i(t))_{i\in I}}$. By an ODE ${\mathcal{D}}$, we mean a collection of $D$ coupled ordinary differential equations with initial conditions $\mathbf{X}^{(k)}_0$:
\begin{align*}
\mathcal{D}: \: \left\lbrace
\begin{array}{ll}
f_i(X_i,\mathbf{X}_{\mathtt{pa}(i)})(t) = 0, \quad X_i^{(k)}(0) = (\mathbf{X}^{(k)}_0)_i, \\ \hfill 0\leq k \leq n_i-1, \quad i \in \mathcal{I},
\end{array}
\right.
\end{align*}
where the $i$th differential equation determines the evolution of the variable $X_i$ in terms of $\mathbf{X}_{\mathtt{pa}(i)}$, where $\mathtt{pa}(i) \subseteq \mathcal{I}$ are the \emph{parents of $i$}, and $X_i$ itself, and where $n_i$ is the order of the highest derivative $X^{(k)}_i$ of $X_i$ that appears in equation $i$. Here, $f_i$ is a functional that can include time-derivatives of its arguments. %In addition to defining the evolution of $X_i$ in terms of $\Xipa$, we can also interpret it as a constraint: ${f_i(X_i,\mathbf{X}_{\mathtt{pa}(i)})(t) = 0 \: \forall t\in \mathbb{R}_{\geq 0}}$ if and only if $X_i$ and $\Xipa$ obey the physical laws of $\mathcal{D}$ that govern the joint dynamics of $X_i$ and $\Xipa$.
% Joris: I just commented out the constraint interpretation, as it's not so obvious how it's relevant to the rest of the story, and it talks only about physical laws (what about ODEs modeling biological / chemical / economical / ... systems?)
We think of the $i$th differential equation as modelling the \emph{causal mechanism} that determines the dynamics of the effect $X_i$ in terms of its direct causes $\mathbf{X}_{\mathtt{pa}(i)}$.

%\Bernhard{could skip this paragraph to save space}
One possible way to write down an ODE is to canonically decompose it into a collection of first order differential equations, such as is done in \cite{MooJanSch13}. We choose to present our ODEs as ``one equation per variable'' rather than splitting up the equations due to complications that would otherwise occur when considering time-dependent interventions (cf.\ Section~\ref{section:ode_interventions}).

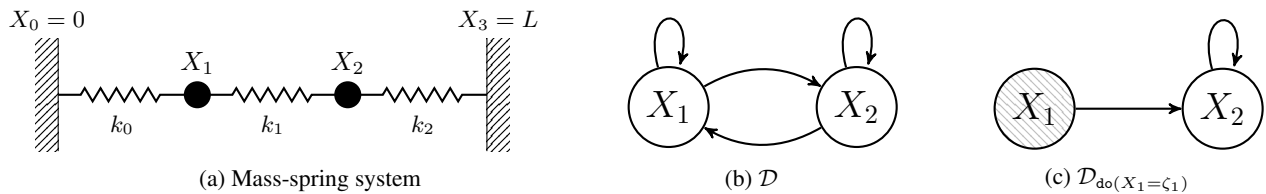
\begin{figure*}
\centering
\begin{subfigure}{0.5\textwidth}
\begin{tikzpicture}[every node/.style={draw,outer sep=0pt,thick}]
\tikzstyle{mass}=[circle,fill=black,minimum width=0.1cm]
\tikzstyle{spring}=[thick,decorate,decoration={zigzag,pre length=0.3cm,post length=0.3cm,segment length=6}]
\tikzstyle{ground}=[fill,pattern=north east lines,draw=none,minimum width=0.75cm,minimum height=0.3cm]

\node (leftwall) [ground, rotate=-90, minimum width=1.5cm,yshift=-2cm,label=left:{\small $X_0=0$}] {};
\draw (leftwall.north east) -- (leftwall.north west);

\node (M1) [mass,xshift=0cm,label={$X_1$}] {};
\node (M2) [mass,xshift=2cm,label={$X_2$}] {};

\draw [spring] (leftwall) -- (M1.west);
\draw [spring] (M1.east) -- (M2.west);

\node (k0) [draw=none, yshift=-0.4cm] at ($(leftwall)!0.5!(M1)$){\small $k_0$};
\node (k1) [draw=none, yshift=-0.4cm] at ($(M1)!0.5!(M2)$){\small $k_1$};

\node (rightwall) [ground, rotate=90, minimum width=1.5cm, yshift = -4cm, label=right:{\small $X_3 = L$}]{};
\draw (rightwall.north west) -- (rightwall.north east);

\draw [spring] (M2.east) -- (rightwall);
\node (k2) [draw=none, yshift=-0.4cm] at ($(M2)!0.5!(rightwall)$){\small $k_2$};
\end{tikzpicture}
\caption{Mass-spring system\label{fig:mass-spring}}
\end{subfigure}
\begin{subfigure}{0.2\textwidth}
\centering
\begin{tikzpicture}[->,>=stealth',auto,node distance=2.5cm,
  thick,main node/.style={circle,draw,font=\sffamily\Large\bfseries}]
\node[main node] (1) {$X	_1$};
\node[main node] (2) [right of=1] {$X_2$};

\draw [->] (1) to [out=30,in=150] (2);
\draw [->] (1) edge [loop above] (1);
\draw [->] (2) to [out=210,in=-30] (1);
\draw [->] (2) edge [loop above] (2);
\end{tikzpicture}
\caption{$\mathcal{D}$\label{fig:subfig:graphical_model_observational}}
\end{subfigure}
\hfill
\begin{subfigure}{0.2\textwidth}
\centering
\begin{tikzpicture}[->,>=stealth',auto,node distance=2.5cm,
  thick,main node/.style={circle,draw,font=\sffamily\Large\bfseries}]
\node[main node, fill,pattern=north west lines, pattern color = lightgray] (1) {$X_1$};
\node[main node] (2) [right of=1] {$X_2$};

\draw [->] (1) to [out=0,in=180] (2);
\draw [->] (2) edge [loop above] (2);
\end{tikzpicture}
\caption{$\mathcal{D}_{\doop(X_1 = \zeta_1)}$\label{fig:subfig:graphical_model_intervened}}
\end{subfigure}
\caption{(a) The mass-spring system of Example \ref{example:3-mass-spring} with $D=2$; 
(b--c) graphs representing the causal structure of the mass-spring system for (b) the observational system, (c) after the intervention on variable $X_1$ described in Example \ref{example:3-mass-spring-intervened}. As a result of the intervention, $X_1$ is not causally influenced by any variable, while the causal mechanism of $X_2$ remains unchanged. \label{fig:graphical_models_mass_spring}}
\end{figure*}

\begin{example}\label{example:3-mass-spring}
Consider a one-dimensional system of $D$ particles of mass ${m_i \: (i=1,\ldots,D)}$ with positions $X_i$ coupled by springs with natural lengths $l_i$ and spring constants $k_i$, where the $i$th spring connects the $i$th and $(i+1)$th masses and the outermost springs have fixed ends (see Figure \ref{fig:mass-spring}). Assume further that the $i$th mass undergoes linear damping with coefficient $b_i$.

Denoting by $\dot{X}_i$ and $\ddot{X}_i$ the first and second time derivatives of $X_i$ respectively, the equation of motion for the $i$th variable is given by
\begin{align*}
m_i \ddot{X}_i(t) =  &k_i[X_{i+1}(t) - X_i(t) - l_i]   \\
&- k_{i-1}[X_i(t) - X_{i-1}(t) - l_{i-1}] - b_i \dot{X}_i(t) 
\end{align*}
where we take ${X_0 = 0}$ and ${X_D=L}$ to be the fixed positions of the end springs. For the case that ${D=2}$, we can write the system of equations as:
\begin{align*}
\mathcal{D}: \left\lbrace
\begin{array}{lll}
0 = m_1 \ddot{X}_1(t) +  b_1 \dot{X}_1(t)  + (k_1 + k_0) X_{1}(t) \\ \hspace{1cm} -  k_1X_{2}(t)  - k_{0} l_{0}+ k_1 l_1 \,, \quad  \\ \\
0 = m_2 \ddot{X}_2(t) +  b_2 \dot{X}_2(t)  + (k_2 + k_{1}) X_{2}(t) \\ \hspace{1cm} -  k_2L  - k_{1} X_{1}(t)  - k_{2} l_1 + k_2 l_2 \,, \\ \\
X_i^{(k)}(0) = (\mathbf{X}^{(k)}_0)_i  \quad k \in \{0,1\}, \: i \in \{1,2\}\,.\\
\end{array}
\right.
\end{align*}
\end{example}

We can represent the functional dependence structure between variables implied by the functions $f_i$ with a graph, in which variables are nodes and arrows point ${X_j \longrightarrow X_i}$ if ${j \in \pa(i)}$. Self loops ${X_i \longrightarrow X_i}$ exist if $X_i^{(k)}$ appears in the expression of $f_i$ for more than one value of $k$. This is illustrated for the system described in Example \ref{example:3-mass-spring} in Figure \ref{fig:subfig:graphical_model_observational}. 

\vspace{-0.15cm}
\section{INTERVENTIONS ON ODES}\label{section:interventions}

We interpret ODEs as \emph{causal} models. In particular, we consider the graph expressing the functional dependence structure to be the causal graph of the system, with an edge between $X_i$ and $X_j$ iff $X_i$ is a direct cause of $X_j$ (in the context of all variables $\mathbf{X}_{\mathcal{I}}$). In this section, we will formalize this causal interpretation by studying interventions on the system.

\vspace{-0.15cm}
\subsection{TIME-DEPENDENT PERFECT INTERVENTIONS}
Usually in the causality literature, by a \emph{perfect intervention} it is meant that a variable is clamped to take a specific given value. The natural analogue of this in the time-dependent case is a perfect intervention that forces a variable to take a particular \emph{trajectory}. That is, given a subset ${I \subseteq \mathcal{I}}$ and a function ${\Zeta_I : \mathbb{R}_{\geq0} \longrightarrow \prod_{i \in I}\mathcal{R}_i}$, we can intervene on the subset of variables $\mathbf{X}_I$ by forcing ${\mathbf{X}_I(t) = \Zeta_I(t) \: \forall t \in \mathbb{R}_{\geq0}}$. Using Pearl's do-calculus notation \citep{Pearl2009} and for brevity omitting the $t$, we write ${\doop(\mathbf{X}_I = \Zeta_I)}$ for this intervention. Such interventions are more general objects than those of the equilibrium or time-independent case, but in the specific case that we restrict ourselves to constant trajectories the two notions coincide.

\vspace{-0.15cm}
\subsection{SETS OF INTERVENTIONS}\label{sec:sets_of_interventions}
Recall that when modelling equilibrating dynamical systems under constant interventions, the set of interventions modelled coincides with the asymptotic behaviour of the system.
We will generalise this relation to non-equilibrating behaviour.

The Dynamic SCMs that we will derive will describe the asymptotic dynamics of the ODE and how they change under different interventions. If we want to model `all possible interventions', then the resulting asymptotic dynamics that can occur are arbitrarily complicated. The idea is to fix a simpler set of interventions and derive an SCM that models only these interventions, resulting in a model that is simpler than the original ODE but still allows us to reason about interventions we are interested in. In the examples in this paper, we restrict ourselves to periodic or quasi-periodic interventions, but the results hold for more general sets of interventions that satisfy the stability definitions presented later.

We need to define some notation to express the sets of interventions and the set of system responses to these interventions that we will model. Since interventions correspond to forcing variables to take some trajectory, we describe notation for defining sets of trajectories: For ${I\subseteq \mathcal{I}}$, let $\intset_I$ be a set of trajectories in ${\prod_{i\in I} \mathcal{R}_i}$. Let ${\intset = \cup_{I \in \mathcal{P}(\mathcal{I})}\intset_I}$ (where $\mathcal{P}(\mathcal{I})$ is the power set of $\mathcal{I}$ i.e., the set of all subsets of $\mathcal{I}$). Thus, an element ${\Zeta_I \in \intset_I}$ is a function ${\mathbb{R}_{\geq 0} \longrightarrow \prod_{i\in I} \mathcal{R}_i}$, and $\intset$ consists of such functions for different $I \subseteq \mathcal{I}$. %\Bernhard{I am not too fond of the notation $\intset$, it looks too much like the integer data type in computer science.}
The main idea is that we want both the interventions and the system responses to be elements of $\intset$; in other words, the set of possible system responses should be large enough to contain all interventions that we would like to model, and in addition, all responses of the system to those interventions. The reader might wonder why we do not simply take the set of \emph{all} possible trajectories, but that set would be so large that it would not be practical for modeling purposes.\footnote{For example, one might want to parameterize the set of trajectories in order to learn the model from data. Without any restriction on the smoothness of the trajectories, the problem of estimating a trajectory from data becomes ill-posed. Secondly, since we would like to identify trajectories that are asymptotically identical in order to focus the modeling efforts on the \emph{asymptotic} behaviour of the system, we will only put a single trajectory into $\intset$ to represent all trajectories that are asymptotically identical to that trajectory, but whose transient dynamics may differ.}

%We alert the reader to the fact that elements of $\intset_\mathcal{I}$ define trajectories for the whole system and will be used in the following sections to define the asymptotic dynamics of the system, in addition to interventions on the entire system. 
Since our goal will be to derive a causal model that describes the relations between components (variables) of the system, we will need the following definition in Section \ref{section:dscm}. 
\begin{definition}
A set of trajectories $\intset$ is \textbf{modular} if, for any ${\{i_1, \ldots, i_n\} = I \subseteq \mathcal{I}}$,
\[ \pmb{\zeta}_I \in \intset \iff \ \zeta_{i_k} \in \intset \quad \forall k \in \{1,\ldots,n\}.\]
\end{definition}
This should be interpreted as saying that admitted trajectories of single variables can be combined arbitrarily into admitted trajectories of the whole system (and \emph{vice versa}, admitted system trajectories can be decomposed into trajectores of individual variables), and in addition, that interventions on each variable can be made independently and combined in any way.\footnote{This is related to notions that have been discussed in the literature under various headings, for instance autonomy and invariance  \citep{Pearl2009}.} This is not to say that all such interventions must be physically possible to implement in practice. Rather, this means that the mathematical model we derive should allow one to \emph{reason} about all such interventions. Not all sets of trajectories $\intset$ are modular; in the following sections we will assume that the sets of trajectories we are considering \textit{are} for the purposes of constructing the Dynamic SCMs. Some examples of trivially modular sets of trajectories are: (i) all static (i.e., time-independent) trajectories, corresponding to \citep{MooJanSch13}; (ii) all continuously-differentiable trajectories that differ asymptotically; (iii) all periodic motions. The latter is the running example in this paper.

%\begin{figure*}
%\centering
%\begin{subfigure}{0.45\textwidth}
%\centering
%\begin{tikzpicture}[->,>=stealth',auto,node distance=2.5cm,
%  thick,main node/.style={circle,draw,font=\sffamily\Large\bfseries}]
%\node[main node] (1) {$X	_1$};
%\node[main node] (2) [right of=1] {$X_2$};
%
%\draw [->] (1) to [out=30,in=150] (2);
%\draw [->] (1) edge [loop above] (1);
%\draw [->] (2) to [out=210,in=-30] (1);
%\draw [->] (2) edge [loop above] (2);
%\end{tikzpicture}
%\caption{$\mathcal{D}$\label{fig:subfig:graphical_model_observational}}
%\end{subfigure}
%\hfill
%\begin{subfigure}{0.45\textwidth}
%\centering
%\begin{tikzpicture}[->,>=stealth',auto,node distance=2.5cm,
%  thick,main node/.style={circle,draw,font=\sffamily\Large\bfseries}]
%\node[main node, fill,pattern=north west lines, pattern color = lightgray] (1) {$X_1$};
%\node[main node] (2) [right of=1] {$X_2$};
%
%\draw [->] (1) to [out=0,in=180] (2);
%\draw [->] (2) edge [loop above] (2);
%\end{tikzpicture}
%\caption{$\mathcal{D}_{\doop(X_1 = \zeta_1)}$\label{fig:subfig:graphical_model_intervened}}
%\end{subfigure}
%\caption{Graphs representing the causal structure of the mass-spring system of Example \ref{example:3-mass-spring} for: (a) the observational system; and (b) after the intervention on variable $X_1$ described in Example \ref{example:3-mass-spring-intervened}. As a result of the intervention, $X_1$ is not causally influenced by any variable, while the causal mechanism of $X_2$ remains unchanged. \label{fig:graphical_models_mass_spring}}
%\end{figure*}

\vspace{-0.15cm}
\subsection{DESCRIBING INTERVENTIONS ON ODEs}\label{section:ode_interventions}

We can realise a perfect intervention by replacing the equations of the intervened variables with new equations that fix them to take the specified trajectories:\footnote{Note that in the intervened ODE, the initial conditions of the intervened variables do not need to be specified explicitly as for the other variables, since they are implied by considering $t=0$.}
\begin{align*}
&\mathcal{D}_{\doop(\mathbf{X}_I = \pmb{\zeta}_I)}: \\
&\left\lbrace
\begin{array}{ll}
f_i(X_i,\mathbf{X}_{\mathtt{pa}(i)})(t) = 0 \,, & X_i^{(k)}(0) = (\mathbf{X}^{(k)}_0)_i \,, \quad \\ 
0\leq k \leq n_i-1 \,, \quad &i \in \mathcal{I}\setminus I  \,, \\ \\ 
X_i(t) - \zeta_i(t) = 0\,, & i \in I \,.
\end{array}
\right.
\end{align*}
This procedure is analogous to the notion of intervention in an SCM. In reality, this corresponds to decoupling the intervened variables from their usual causal mechanism by forcing them to take a particular value, while leaving the non-intervened variables' causal mechanisms unaffected.

Perfect interventions will not generally be realisable in the real world. In practice, an intervention on a variable would correspond to altering the differential equation governing its evolution by adding extra forcing terms; perfect interventions could be realised by adding forcing terms that push the variable towards its target value at each instant in time, and considering the limit as these forcing terms become infinitely strong so as to dominate the usual causal mechanism determining the evolution of the variable.

\begin{example}[continued]\label{example:3-mass-spring-intervened}
Consider the mass-spring system described in Example \ref{example:3-mass-spring}. If we were to intervene on the system to force the mass $X_1$ to undergo simple harmonic motion, we could express this as a change to the system of differential equations as:
\begin{align*}
&\mathcal{D}_{\doop(X_1(t) = l_1 + A\cos(\omega t))}: \\
&\left\lbrace
\begin{array}{l}
0 = X_1(t)- l_1 - A \cos(\omega t) \,,  \\ \\
0 = m_2\ddot{X}_2(t) +  b_2\dot{X}_2(t)  + (k_2 + k_{1}) X_{2}(t) \\ \hfill -  k_2L  - k_{1} X_{1}(t)  - k_{2} l_1 + k_2 l_2 \,, \\ \\
X_2^{(k)}(0) = (\mathbf{X}^{(k)}_0)_2  \quad k \in \{0,1\}.\\
\end{array}
\right.
\end{align*}
\end{example}
This induces a change to the graphical description of the causal relationships between the variables. We break any incoming arrows to any intervened variable, including self loops, as the intervened variables are no longer causally influenced by any other variable in the system. See Figure \ref{fig:subfig:graphical_model_intervened} for the graph corresponding to the intervened ODE in Example \ref{example:3-mass-spring-intervened}.

%\begin{itemize}
%\item Given the path of a variable, we can realise this as an intervention on the ODE by replacing the variable's equation
%\item Define sets of interventions and observe that these are sets of paths.
%\item Set up intervention notation for rest of paper?
%\end{itemize}

\vspace{-0.15cm}
\section{DYNAMIC STABILITY}\label{section:dynamic-stability}

%\begin{itemize}
%\item Define dyn.
%\item Give dynamic stability definitions (of a system, wrt an intervention, wrt a set of interventions)
%\item Interventional closure
%\end{itemize}

%In the original paper, the dynamical systems considered had the property that the asymptotic dynamics were always constant. By considering the $t \longrightarrow \infty$ limit in which any complex but transient dynamical behaviour would have decayed, the authors were able to derive SCMs to reason about asymptotic behaviour of the systems under interventions. Crucially, the interventions modelled had the same functional form as the assumed asymptotic dynamics of the systems - they were constant.
%
%The key insight leading to the approach we take in this paper is that we can relax the assumption that the asymptotic dynamics must be constant; if they can be expressed 

A crucial assumption of \cite{MooJanSch13} was that the systems considered were \emph{stable} in the sense that they would converge to unique stable equilibria (if necessary, also after performing a constant intervention). This made them amenable to study by considering the ${t \longrightarrow \infty}$ limit in which any complex but transient dynamical behaviour would have decayed. The SCMs derived would allow one to reason about the asymptotic equilibrium states of the systems after interventions. Since we want to consider non-constant asymptotic dynamics, this is not a notion of stability that is fit	 for our purposes.

Instead, we define our stability with reference to a set of trajectories. We will use $\intset_\mathcal{I}$ for this purpose. Recall that elements of $\intset_\mathcal{I}$ are trajectories for all variables in the system. To be totally explicit, we can think of an element ${\pmb{\eta} \in \intset_\mathcal{I}}$ as a function  
\begin{align*}
\pmb{\eta}:  \quad \mathbb{R}_{\geq0} & \longrightarrow  \mathcal{R}_\mathcal{I} \\
 t & \mapsto  (\eta_1(t),\eta_2(t),\ldots,\eta_D(t))
\end{align*}
where $\eta_i(t) \in \mathcal{R}_i$ is the state of the $i$th variable $X_i$ at time $t$. Note that $\intset_\mathcal{I}$ is not a single fixed set, independent of the situation we are considering. We can choose $\intset_\mathcal{I}$ depending on the ODE $\mathcal{D}$ under consideration, and the interventions that we may wish to make on it. 

%We will identify any two trajectories that have the same asymptotic behavior: for $\pmb{\eta},\pmb{\zeta}\in\intset_\mathcal{I}$, $\pmb{\eta} \sim \pmb{\zeta} \iff \lim_{t\to\infty} |\pmb{\eta} - \pmb{\zeta}| = 0$.
Informally, stability in this paper means that the asymptotic dynamics of the dynamical system converge to a unique element of $\intset_\mathcal{I}$, independent of initial condition. If $\intset_\mathcal{I}$ is in some sense simple, we can simply characterise the asymptotic dynamics of the system under study. The following definitions of stability extend those of \cite{MooJanSch13} to allow for non-constant trajectories in $\intset_{\mathcal{I}}$, and coincide with them in the case that $\intset_{\mathcal{I}}$ consists of all constant trajectories in $\mathcal{R}_\mathcal{I}$.

%\begin{example}
%A more expressive set for $\dynset$ might be the (length $D$) cartesian product of all 1-dimensional paths with constant fourier decomposition. \textcolor{red}{Not sure how best to describe this concisely. I mean that we take the set of 1-dimensional paths with constant fourier decomposition, and allow each of the $D$ variables to have path in this set independently of one another}
%\end{example}

%\begin{definition}\label{def:single_stable}
%  The ODE $\mathcal{D}$ is \textbf{dynamically stable with reference to} $\intset_\mathcal{I}$ if there exists a unique $\pmb{\eta}_\emptyset \in \intset_\mathcal{I}$ such that:
%%
%  \begin{compactitem}
%  \item there exist initial conditions $\mathbf{X}^{(k)}_0$ such that ${\mathbf{X}_{\mathcal{I}}(t) = \Eta_\emptyset(t) \: \forall t}$ is a solution to $\mathcal{D}$;
%  \item for all initial conditions $\mathbf{X}^{(k)}_0$, ${\mathbf{X}_\mathcal{I}(t) \rightarrow \pmb{\eta}_\emptyset(t)}$ as $t \rightarrow \infty$.
%  \end{compactitem}
%%
%\end{definition}
\begin{definition}\label{def:single_stable}
  The ODE $\mathcal{D}$ is \textbf{dynamically stable with reference to} $\intset_\mathcal{I}$ if there exists a unique $\pmb{\eta}_\emptyset \in \intset_\mathcal{I}$ such that ${\mathbf{X}_{\mathcal{I}}(t) = \Eta_\emptyset(t) \: \forall t}$ is a solution to $\mathcal{D}$ and that for any initial condition, the solution ${\mathbf{X}_\mathcal{I}(t) \rightarrow \pmb{\eta}_\emptyset(t)}$ as $t \rightarrow \infty$.\footnote{The convergence we refer to here is the usual asymptotic convergence of real-valued functions, i.e., for $f : [0,\infty) \to \mathbb{R}^d$, $g : [0,\infty) \to \mathbb{R}^d$ we have that $f \to g$ iff for every $\epsilon > 0$ there is a $T \in [0,\infty)$ such that $|f(t) - g(t)| < \epsilon$ for all $t \in [T,\infty)$.}
\end{definition}

%\begin{definition}\label{def:single_stable2}
%The ODE $\mathcal{D}$ is \textbf{dynamically stable with reference to} $\intset_\mathcal{I}$ if 
%\begin{itemize}
%  \item there exists a $\pmb{\eta}_\emptyset \in \intset_\mathcal{I}$ such that ${\mathbf{X}_{\mathcal{I}}(t) = \Eta_\emptyset(t) \: \forall t}$ is a solution to $\mathcal{D}$, and for any initial condition, the solution ${\mathbf{X}_\mathcal{I}(t) \rightarrow \pmb{\eta}_\emptyset(t)}$ as $t \rightarrow \infty$;
%  \item $\pmb{\eta}_\emptyset \in \intset_\mathcal{I}$ and $\pmb{\eta}_\emptyset' \in \intset_\mathcal{I}$ such that $\pmb{\eta}_\emptyset \ne \pmb{\eta}_\emptyset'$ implies $\pmb{\eta}_\emptyset(t) - \pmb{\eta}_\emptyset'(t) \nrightarrow 0$ for $t \to \infty$.
%\end{itemize}
%\end{definition}

We use a subscript $\emptyset$ to emphasise that $\Eta_\emptyset$ describes the asymptotic dynamics of $\mathcal{D}$ without any intervention. Observe that $\intset_\mathcal{I}$ could consist of the single element $\Eta_\emptyset$ in this case. The requirement that this hold for all initial conditions can be relaxed to hold for all initial conditions except on a set of measure zero, but that would mean that the proofs later on require some more technical details. For the purpose of exposition, we stick to this simpler case.

\begin{example}\label{example:trivial-non-constant}
Consider a single mass on a spring that is undergoing simple periodic forcing and is underdamped. Such a system could be expressed as a single (parent-less) variable with ODE description:
\begin{align*}
\mathcal{D}: \left\lbrace
\begin{array}{ll}
m \ddot{X_1}(t) + b \dot{X_1}(t) + k(X_1(t)-l) \\ \hfill= F \cos(\omega t + \phi) \,, \\ \\ \hfill X_1^{(k)}(0) = (X^{(k)}_0)  \quad k \in \{0,1\} \,.
\end{array}
\right.
\end{align*}

The solution to this differential equation is
\begin{equation}\label{eqn:mass-spring-solution}
X_1(t) = r(t) + l +  A \cos(\omega t + \phi') %\tag{$*$}
\end{equation}
where $r(t)$ decays exponentially quickly (and is dependent on the initial conditions) and $A$ and $\phi'$ depend on the parameters of the equation of motion (but not on the initial conditions).

Therefore such a system would be dynamically stable with reference to (for example)
\[\intset_\mathcal{I} = \{l + A\cos(\omega t + \phi') : A\in \mathbb{R}, \: \phi' \in [0,2\pi) \}. \]
\end{example}

\begin{remark}
We use a subscript $\pmb{\zeta}_I$ to emphasise that $\Eta_{\pmb{\zeta}_I}$ describes the asymptotic dynamics of $\mathcal{D}$ after performing the intervention $\doop(\mathbf{X}_I = \pmb{\zeta}_I)$. 
Observe that $\intset_{\mathcal{I}}$ could consist only of the single element $\Eta_{\pmb{\zeta}_{I}}$ and the above definition would be satisfied. But then the original ODE wouldn't be dynamically stable with reference to $\intset_{\mathcal{I}}$, nor would other intervened versions of $\mathcal{D}$. This motivates the following definition, extending dynamic stability to sets of intervened systems.
\end{remark}

\begin{definition}\label{def:intset_stable}

%For any $\pmb{\zeta} \in \intset$, we write $\mathtt{do}(\mathbf{X} = \pmb{\zeta})$ for the intervention represented by $\pmb{\zeta}$, where it is understood that we only intervene on the variables for which $\pmb{\zeta}$ defines a path, and leave the equations governing the evolution of the other variables unchanged.

Let $\pathset$ be a set of trajectories. We say that the pair $(\mathcal{D}, \pathset)$ is \textbf{dynamically stable with reference to} $\intset_\mathcal{I}$ if, for any $\pmb{\zeta}_I\in \pathset$\,,  $\mathcal{D}_{\mathtt{do}(\mathbf{X}_I=\pmb{\zeta}_I)} $ is dynamically stable with reference to $\intset_\mathcal{I}$.
\end{definition}

%\begin{example}
%%Let us return again to the mass-spring chain of Example \ref{example:3-mass-spring}. Suppose we take $\intset$ to be the set of all constant interventions, and $\dynset$ to be the set of all constant paths. Then $(\mathcal{D},\intset)$ is dynamically stable with reference to $\dynset$. 
%
%Suppose that $\mathcal{D}$ is \emph{stable} in the sense of [Mooij et al]. Then if $\intset$ and $\dynset$ both consist of all constant paths, then $(\mathcal{D}, \intset)$ will be dynamically stable with reference to $\dynset$.
%\end{example}

\begin{figure*}
\centering
\begin{subfigure}{0.48\textwidth}
\centering
\includegraphics[scale=0.35]{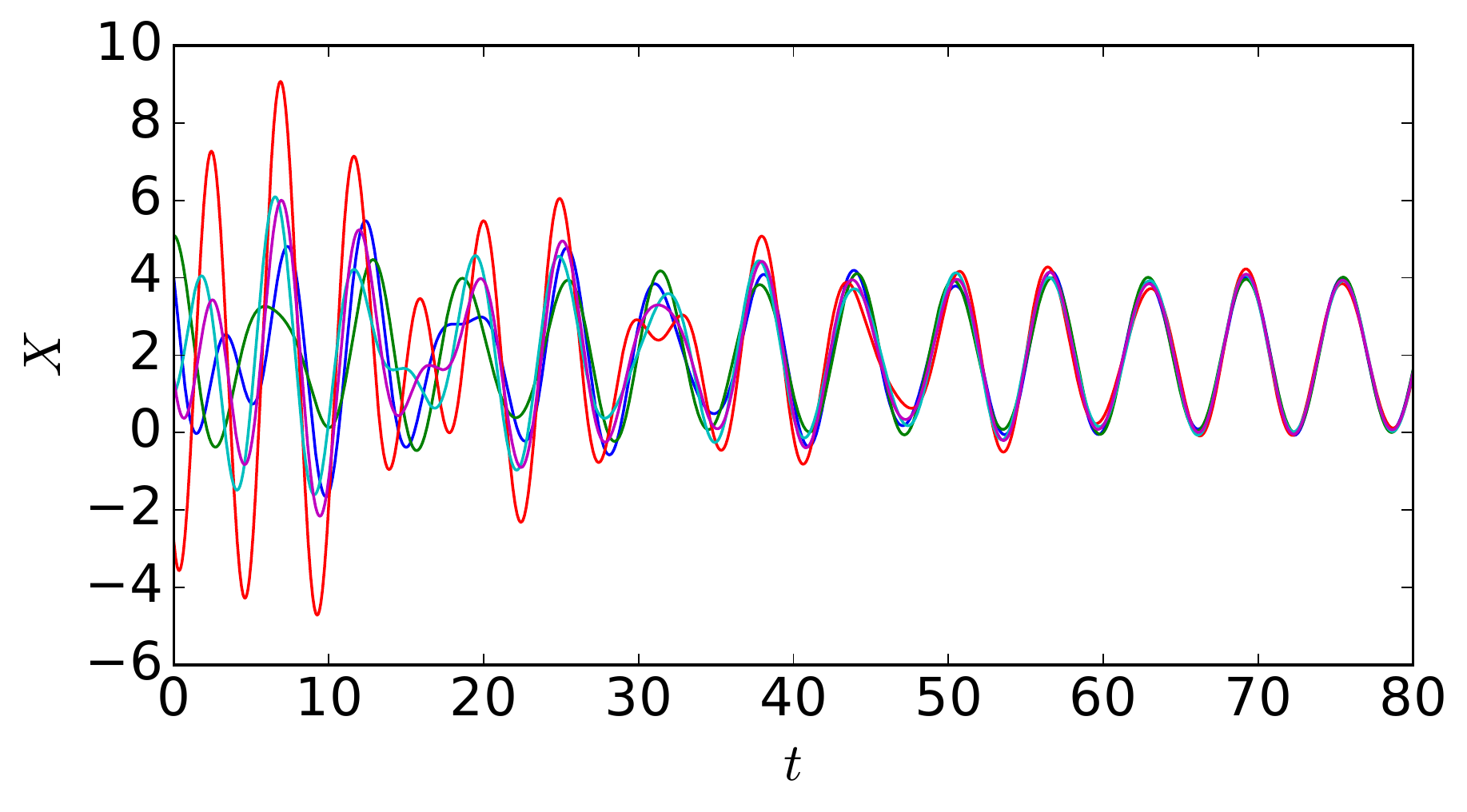}
\caption{\label{subfig:decay_shm_1}}
\end{subfigure}
\begin{subfigure}{0.48\textwidth}
\centering
\includegraphics[scale=0.35]{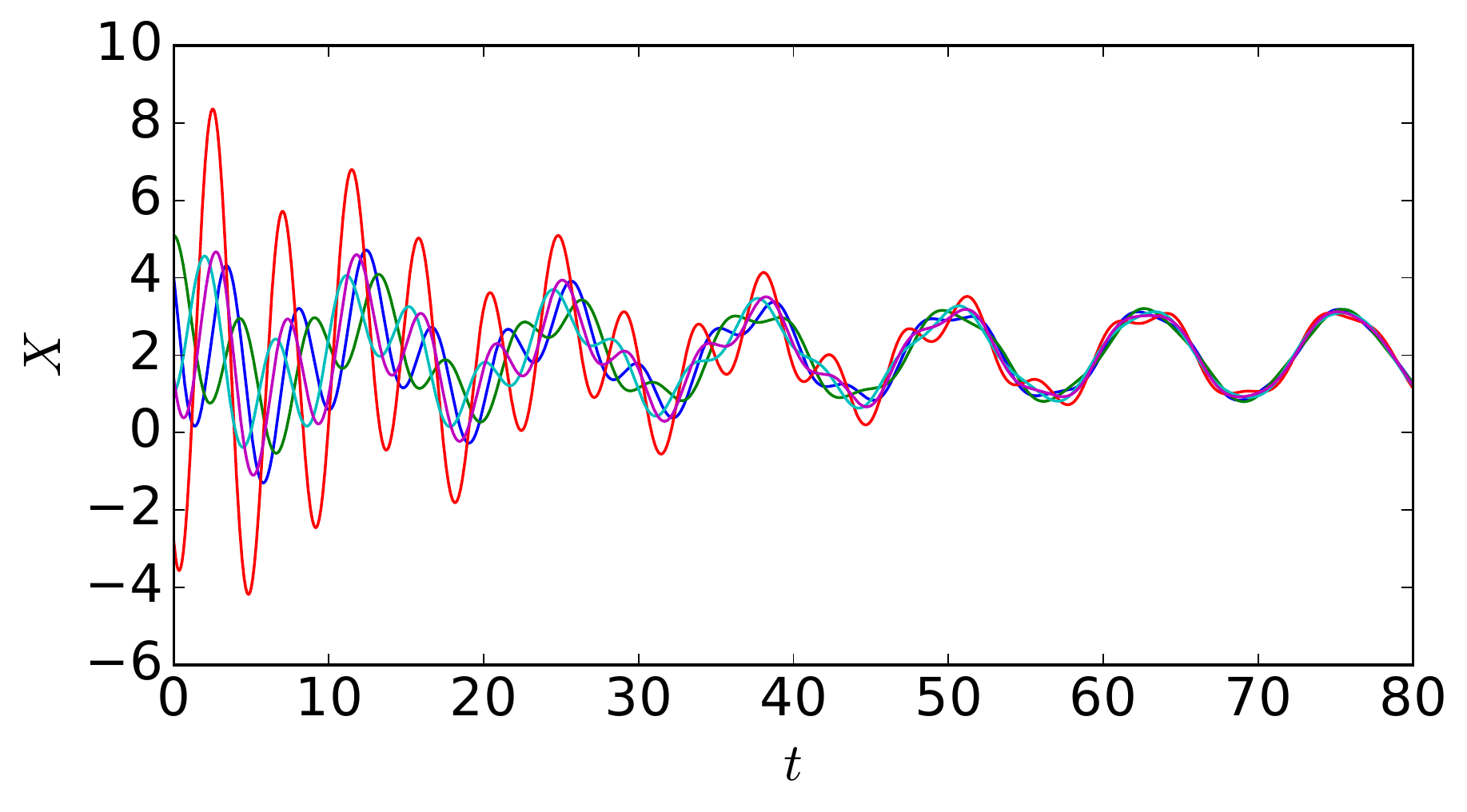}
\caption{\label{subfig:decay_shm_2}}
\end{subfigure}
  \caption{Simulations from the forced simple harmonic oscillator in Example \ref{example:trivial-non-constant} showing the evolution of $X_1$ with different initial conditions for different forcing terms (interventions on $X_2$). The parameters used were $m=1, k=1, l=2, F=2 , b=0.1$, with (a) $\omega = 3$ and (b) $\omega=2$. Dynamic stability means that asymptotic dynamics are independent of initial conditions, and the purpose of the DSCM is to quantify how the asymptotic dynamics change under intervention.\label{fig:decay_shm}}
\end{figure*}

\begin{contexample}
%Consider again the single-mass-spring system.
% of Example \ref{example:trivial-non-constant}. 
Suppose we are interested in modelling the effect of changing the forcing term, either in amplitude, phase or frequency. We introduce a second variable $X_2$ to model the forcing term:
\begin{align*}
\mathcal{D}: \left\lbrace
\begin{array}{lll}
0 &= f_1(X_1,X_2)(t)  \\ & =  m \ddot{X}_1(t) + b \dot{X}_1(t) + k(X_1(t)-l)  -  X_2(t) \,,  \\ \\ 
0 &= f_2(X_2) (t)  \\  & =  X_2(t)  - F_0 \cos(\omega_0 t + \phi_0) \,, \\
\\
&X_1^{(k)}(0) = (\mathbf{X}^{(k)}_0)_1\,,  \quad k \in \{0,1\}\, .
\end{array}
\right.
\end{align*}

If we want to change the forcing term that we apply to the mass, we can interpret this as performing an intervention on $X_2$. We could represent this using the notation we have developed as 
\begin{align*} \intset_{\{2\}} = \{ \zeta_2(t) =  F_2 \cos (\omega t + \phi_2) : \\
  \: F_2, \omega \in \mathbb{R}, \: \phi_2 \in [0,2\pi) \}. 
\end{align*}
For any intervention $\zeta_2 \in \intset_{\{2\}}$, the dynamics of $X_1$ in $\mathcal{D}_{\doop(X_2 = \zeta_2)}$ will be of the form (\ref{eqn:mass-spring-solution}). Therefore $(\mathcal{D}, \intset_{\{2\}})$ will be dynamically stable with reference to 
\begin{align*}
 \intset_\mathcal{I} = \Big{\{} \pmb{\zeta}(t) =  (l + F_1 \cos (\omega t + \phi_1),  F_2 \cos (\omega t + \phi_2)) \\ : \: F_1,F_2, \omega \in \mathbb{R}, \: \phi_1,\phi_2 \in [0,2\pi) \Big{\}}.
 \end{align*}
\end{contexample}
The independence of initial conditions for Example \ref{example:trivial-non-constant} is illustrated in Figure \ref{fig:decay_shm}.

Note that if ${(\mathcal{D},\pathset)}$ is dynamically stable with reference to ${\intset_\mathcal{I}}$, and ${\intset_\mathcal{I}' \supseteq \intset_\mathcal{I}}$ is a larger set of trajectories that still satisfies the uniqueness condition in the definition of dynamic stability,\footnote{Namely: $ \forall \Zeta_I \in \pathset,\: \exists!\, \Eta_{\pmb{\zeta}_I}\in \intset_\mathcal{I}' $ such that under $\mathcal{D}_{\mathtt{do}(\mathbf{X}_I=\pmb{\zeta}_I)}$ and for any initial condition, $X_\mathcal{I}(t) \rightarrow \Eta_{\pmb{\zeta}_I}(t)$ as $t\rightarrow \infty$. Assuming that $(\mathcal{D},\pathset)$ is dynamically stable with reference to $\intset_\mathcal{I}$, a sufficient condition for this is that none of the elements in $\intset_\mathcal{I}'\setminus\intset_\mathcal{I}$ are asymptotically equal to any of the elements of $\intset_\mathcal{I}$. That is: $\forall \Zeta \in \intset_\mathcal{I},\, \forall\Zeta' \in \intset_{\mathcal{I}}'\setminus \intset_\mathcal{I} $,  ${\Zeta(t) \nrightarrow \Zeta'(t)}$ as ${t \rightarrow \infty}$\,.} then ${(\mathcal{D},\pathset)}$ is dynamically stable with reference to $\intset_\mathcal{I}'$.

\vspace{-0.15cm}
\section{DYNAMIC STRUCTURAL CAUSAL MODELS}\label{section:dscm}

A deterministic SCM $\mathcal{M}$ is a collection of structural equations, the $i$th of which defines the value of variable $X_i$ in terms of its parents. We extend this to the case that our variables do not take fixed values but rather represent entire trajectories.

\begin{definition} 
  Let $\intset=\bigcup_{I\subseteq\mathcal{I}} \intset_I$ be a modular set of trajectories, where $\intset_I\subseteq \mathcal{R}_I^{\mathbb{R}_{\geq 0}}$.
A deterministic Dynamic Structural Causal Model (DSCM) on the time-indexed variables $\mathbf{X}_\mathcal{I}$ taking values in $\intset$ is a collection of \emph{structural equations}
\begin{align*}
\mathcal{M}: \left\lbrace
\begin{array}{ll}
X_i = F_i(\mathbf{X}_{\pa(i)}) & i \in \mathcal{I} \,,
\end{array}
\right.
\end{align*}
where ${\pa(i) \subseteq \mathcal{I}\setminus \{i\}}$ and each $F_i$ is a map ${\intset_{\pa(i)}\longrightarrow \intset_i}$ that gives the trajectory of an effect variable in terms of the trajectories of its direct causes.
\end{definition}

The point of this paper is to show that, subject to restrictions on $\mathcal{D}$ and $\intset$, we can derive a DSCM that allows us to reason about the effect on the asymptotic dynamics of interventions using trajectories in $\intset$. `Traditional' deterministic SCMs arise as a special case, where all trajectories are constant over time.

%`Traditional' deterministic SCMs and DSCMs are identical except for the fact that the variables in an SCM usually take value in a discrete set or a finite dimensional space such as $\mathbb{R}^n$, whilst variables in a DSCM are trajectories in such sets (and as such are infinite dimensional).

In an ODE, the equations $f_i$ determine the causal relationship between the variable $X_i(t)$ and its parents $\mathbf{X}_{\pa(i)}(t)$ \emph{at each instant} in time. In contrast, we think of the function $F_i$ of the DSCM as a causal mechanism that determines the entire trajectory of $X_i$ in terms of the trajectories of the variables $\mathbf{X}_{\pa(i)}$, integrating over the instantaneous causal effects over all time. In the case that $\intset$ consists of constant trajectories (and thus the instantaneous causal effects are constant over time), a DSCM reduces to a traditional deterministic SCM. 

The rest of this section is laid out as follows. In Section~\ref{section:scm_interventions} we define what it means to make an intervention in a DSCM. In Section~\ref{section:struc_eqns_and_scm} we show how, subject to certain conditions, a DSCM can be derived from a pair ${(\mathcal{D},\intset)}$. The procedure for doing this relies on intervening on all but one variable at a time. In Section~\ref{section:solutions-to-dscm}, Theorem~\ref{theorem:same-solutions} states that the DSCM thus derived is capable of modelling the effect of intervening on arbitrary subsets of variables, even though it was constructed by considering the case that we consider interventions on exactly ${D-1}$ variables. Theorem~\ref{theorem:commuting-diagram} and Corollary~\ref{corr:double-commuting-diagram} in Section~\ref{section:causal-reasoning-preserved} prove that the notions of intervention in ODE and the derived DSCM coincide. Collectively, these theorems tell us that we can derive a DSCM that allows us to reason about the effects of interventions on the asymptotic dynamics of the ODE. Proofs of these theorems are provided in Section~\ref{supp:proofs} of the Supplementary Material.

\vspace{-0.15cm}
\subsection{INTERVENTIONS IN A DSCM}\label{section:scm_interventions}
%DSCMs inherit the same calculus for modelling interventions as SCMs. 
Interventions in (D)SCMs are realized by replacing the structural equations of the intervened variables. Given $\Zeta_I \in\intset_I$ for some $I \subseteq \mathcal{I}$, the intervened DSCM $\mathcal{M}_{\doop(\mathbf{X}_I = \Zeta_I)}$ can be written:
\begin{align*}
\mathcal{M}_{\doop(\mathbf{X}_I = \Zeta_I)}: \left\lbrace
\begin{array}{lll}
X_i &= F_i(\mathbf{X}_{\pa(i)}) & i \in \mathcal{I}\setminus I \,,\\
X_i &= \zeta_i & i \in I \,.\\
\end{array}
\right.
\end{align*}
The causal mechanisms determining the non-intervened variables are unaffected, so their structural equations remain the same. The intervened variables are decoupled from their usual causal mechanisms and are forced to take the specified trajectory.

\vspace{-0.15cm}
\subsection{DERIVING DSCMs FROM ODEs}\label{section:struc_eqns_and_scm}

In order to derive a DSCM from an ODE, we require the following consistency property between the asymptotic dynamics of the ODE and the set of interventions. 

\begin{definition}[Structural dynamic stability]
Let $\intset$ be modular.  The pair $(\mathcal{D},\intset)$ is \textbf{structurally dynamically stable} if $(\mathcal{D},\intset_{\mathcal{I}\setminus \{ i \} })$ is dynamically stable with reference to $\intset_{\mathcal{I}}$ for all i.
\end{definition}

This means that for any intervention trajectory ${\Zeta_{\mathcal{I}\setminus \{ i \} } \in \intset_{\mathcal{I}\setminus \{ i \} }}$, the asymptotic dynamics of the intervened ODE ${\mathcal{D}_{\doop(\mathbf{X}_{\mathcal{I}\setminus \{ i \} } = \Zeta_{\mathcal{I}\setminus \{ i \} })}}$ are expressible uniquely as an element of $\intset_\mathcal{I}$. Since $\intset$ is modular, the asymptotic dynamics of the non-intervened variable can be realised as the trajectory $\zeta_{i} \in \intset_{i}$, and thus $\intset$ is rich enough to allow us to make an intervention which forces the non-intervened variable to take this trajectory. This is a crucial property that allows the construction of the structural equations. In the particular case that $\intset$ consists of all constant trajectories, structural dynamic stability means that after any intervention on all-but-one-variable, the non-intervened variable settles to a unique equilibrium. In the language of \cite{MooJanSch13}, this would imply that the ODE is \emph{structurally stable}.

It should be noted that $(\mathcal{D},\intset)$ being structurally dynamically stable is a strong assumption in general. If $\intset$ is too small,\footnote{For example, if $\intset$ is not modular or represents interventions on only a subset of the variables.} then it may be possible to find a larger set $\intset' \supset \intset$ such that  $(\mathcal{D},\intset')$ \emph{is} structurally dynamically stable. The procedure described in this section describes how to derive a DSCM capable of modelling all interventions in $\intset'$, which can thus be used to model interventions in $\intset$. 

%\Bernhard{I am not sure if the following fits here --- maybe too detailed for the conclusions of the paper? Shorten it, or move elsewhere?}
%In general, we suppose that one would present with\Bernhard{I don't understand this} an ODE $\mathcal{D}$ and a set of interventions $\intset'$ to be modelled, but that in general $(\mathcal{D},\intset')$ would not be interventionally closed. For example, $\intset'$ might only represent interventions on a subset of all of the variables. One would need to find a set $\intset \supseteq \intset'$ such that $(\mathcal{D},\intset)$ is interventionally closed, and then one would be able to model the effects of interventions in $\intset'$ (as well as those in the larger set $\intset$).

%\subsubsection*{Constructing the Structural Equations}
Henceforth, we use the notation $I_i = \mathcal{I}\setminus \{i\}$ for brevity.
Suppose that $(\mathcal{D},\intset)$ is structurally dynamically stable. We can {\bf derive structural equations} ${F_i : \intset_{\pa(i)} \longrightarrow \intset_i }$ to describe the asymptotic dynamics of children variables as functions of their parents as follows. Pick $i\in \mathcal{I}$. The variable $X_i$ has parents $\mathbf{X}_{\pa(i)}$. Since $\intset$ is modular, for any configuration of parent dynamics $\Eta_{\pa(i)} \in \intset_{\pa(i)}$ there exists $\Zeta_{I_i} \in \intset_{I_i}$ such that $(\Zeta_{I_i})_{\pa(i)} = \Eta_{\pa(i)}$.

By structural dynamic stability, the system $\mathcal{D}_{\doop(\mathbf{X}_{I_i} =  \Zeta_{I_i})}$ has asymptotic dynamics specified by a unique element $\Eta \in \intset_\mathcal{I}$, which in turn defines a unique element $\eta_i \in \intset_i$ specifying the asymptotic dynamics of variable $X_i$ since $\intset$ is modular. 

\begin{theorem}\label{theorem:structural-equations-well-defined}
Suppose that $(\mathcal{D},\intset)$ is structurally dynamically stable. Then the functions
\[
  F_i : \intset_{\pa(i)} \to \intset_i :\Eta_{\pa(i)} \mapsto \eta_i
\]
constructed as above are well-defined.
\end{theorem}
%For the proof, see Section \ref{supp:theorem1proof} in the Supplementary Material.

%\Bernhard{This is stated after each theorem --- if we want to save space, we could state once that all theorems are proved in the suppl.}

Given the structurally dynamically stable pair $(\mathcal{D},\intset)$ we define the derived DSCM
\begin{align*}
\mathcal{M}_\mathcal{D}: \left\lbrace
\begin{array}{ll}
X_i = F_i(\mathbf{X}_{\pa(i)}) & i \in \mathcal{I} \,,
\end{array}
\right.
\end{align*}
where the $F_i: \intset_{\pa(i)} \to \intset_i $ are defined as above. Note that structural dynamic stability was a crucial property that ensured $F_i(\intset_{\pa(i)}) \subseteq \intset_i$. If $(\mathcal{D},\intset)$ is not structurally dynamically stable, we cannot build structural equations in this way. 

\begin{figure*}[h!]
\centering
\begin{tikzpicture}[->,>=stealth',auto,node distance=2.5cm,
  thin,main node/.style={rectangle,draw,minimum width=3.3cm, minimum height=1.3cm}]
\node[main node,align=center] (1)  {ODE \\ $\mathcal{D}$};
\node[main node,align=center] (2) [right of=1, xshift=2.8cm] {Intervened ODE \\ $\mathcal{D}_{\doop(\mathbf{X}_I = \Zeta_I)}$};
\node[main node,align=center] (3) [below of=1, yshift=-0.5cm] {DSCM \\ $\mathcal{M}_\mathcal{D}$};
\node[main node,align=center] (4) [below of=2, yshift=-0.5cm] {Intervened DSCM \\ $\mathcal{M}_{\mathcal{D}_{\doop(\mathbf{X}_I = \Zeta_I)}}$};
\node[main node,align=center] (5) [right of=2, xshift=2.8cm] {Intervened ODE \\ $\mathcal{D}_{\doop(\mathbf{X}_I = \Zeta_I, \mathbf{X}_J = \Zeta_J)}$};
\node[main node,align=center] (6) [below of=5, yshift=-0.5cm] {Intervened DSCM \\ $\mathcal{M}_{\mathcal{D}_{\doop(\mathbf{X}_I = \Zeta_I, \mathbf{X}_J = \Zeta_J)}}$};
\draw [|->,  shorten <=2pt, shorten >=2pt] (1.south) to  (3.north) ;
\draw [|->,  shorten <=2pt, shorten >=2pt] (1.east) to  (2.west);
\draw [|->,  shorten <=2pt, shorten >=2pt] (2.south) to  (4.north);
\draw [|->,  shorten <=2pt, shorten >=2pt] (3.east) to  (4.west);
\draw [|->,  shorten <=2pt, shorten >=2pt] (5.south) to  (6.north) ;
\draw [|->,  shorten <=2pt, shorten >=2pt] (2.east) to  (5.west);
\draw [|->,  shorten <=2pt, shorten >=2pt] (4.east) to  (6.west);

\node [draw=none, above=0.2cm] (a) at ($(1)!0.5!(2)$) {Sec.~\ref{section:ode_interventions}};
\node [draw=none, above=0.2cm] (a) at ($(3)!0.5!(4)$) {Sec.~\ref{section:scm_interventions}};
\node [draw=none, right=0.2cm] (a) at ($(1)!0.5!(3)$) {Sec.~\ref{section:struc_eqns_and_scm}};
\node [draw=none, right=0.2cm] (a) at ($(2)!0.5!(4)$) {Sec.~\ref{section:struc_eqns_and_scm}};
\node [draw=none, above=0.2cm] (a) at ($(2)!0.5!(5)$) {Sec.~\ref{section:ode_interventions}};
\node [draw=none, above=0.2cm] (a) at ($(4)!0.5!(6)$) {Sec.~\ref{section:scm_interventions}};
\node [draw=none, right=0.2cm] (a) at ($(5)!0.5!(6)$) {Sec.~\ref{section:struc_eqns_and_scm}};
\end{tikzpicture}
  \caption{Top-to-bottom arrows: Theorems \ref{theorem:structural-equations-well-defined} and \ref{theorem:same-solutions} together state that if $(\mathcal{D},\intset)$ is structurally dynamically stable then we can construct a DSCM to describe the asymptotic behaviour of $\mathcal{D}$ under different interventions in the set $\intset$. Left-to-right arrows: Both ODEs and DSCMs are equipped with notions of intervention. Theorem \ref{theorem:commuting-diagram} and Corollary~\ref{corr:double-commuting-diagram} say that these two notions of intervention coincide, and thus the diagram commutes.  \label{fig:commuting_diagram}}
\end{figure*}
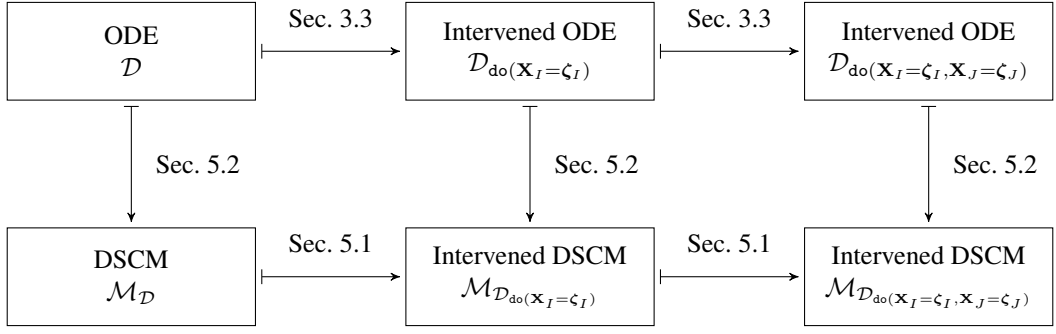

We provide next an example of a DSCM for the mass-spring system of Example \ref{example:3-mass-spring} with $D=2$. 
The derivation of this for the general case of arbitrarily many masses is included in the Supplementary Material.

\begin{example}\label{example:dscm}
	Consider the system $\mathcal{D}$ governed by the differential equation of Example \ref{example:3-mass-spring} with $D=2$. Let $\dynset_{\{1,2\}}$ be the modular set of trajectories with 
	\begin{align*}
	\dynset_{\{i\}} = \Bigg\lbrace  & \sum_{j=1}^\infty A_i^j \cos(\omega_i^j t + \phi_i^j) \: : \\
	& w_i^j, \phi_i^j, A_i^j \in \mathbb{R}, \sum_{j=1}^\infty |A_i^j| < \infty\Bigg\rbrace
	\end{align*}
  for $i=1,2$,
	where for each $i$ it holds that $\sum_{j=1}^\infty |A_i^j| < \infty$ (so that the series is absolutely convergent). 
	Then $(\mathcal{D}, \dynset_{\{1,2\}})$ is structurally dynamically stable and admits the following DSCM.
	\begin{align*}
	\mathcal{M}: \left\lbrace
	\begin{array}{lll}
	X_1 &= F_1(X_2) \\
	X_2 &= F_2(X_1)\\
	\end{array}
	\right.
	\end{align*}
	where, writing $C_1^j = [k_1 + k_{2} - m_1(\omega_{2}^j)^2]^2$ and $C_2^j = [k_1 + k_{2} - m_2(\omega_{1}^j)^2]^2$, the functionals $F_1$ and $F_2$ are given by Equations \ref{eqn:structural_equations_1} and \ref{eqn:structural_equations_2} overleaf.	
\end{example}

\begin{figure*}
	\begin{align}
	\resizebox{0.85\textwidth}{!}{\parbox{\textwidth}{$$
			F_1 \left(\sum_{j=1}^\infty A_{2}^j \cos(\omega_{2}^j t + \phi_{2}^j) \right)  =\frac{- k_{1}l_1}{k_1 + k_{0}} +\sum_{j=1}^\infty  \frac{k_{1}A_{2}^j}{\sqrt{C_1^j + b_1m_1(\omega_{2}^j)^2}} \cos\left(\omega_{2}^j  t + \phi_{2}^j  -  \arctan\left[\frac{b_1\omega_{2}^j }{C_1^j}\right]\right) $$
	}} \label{eqn:structural_equations_1}\\
	\resizebox{0.85\textwidth}{!}{\parbox{\textwidth}{$$
			F_2 \left(\sum_{j=1}^\infty A_{1}^j \cos(\omega_{1}^j t + \phi_{1}^j) \right)  =\frac{k_{1}l_1 - k_2l_2}{k_1 + k_{2}} + \frac{k_2L}{k_2 + k_3} +\sum_{j=1}^\infty  \frac{k_{1}A_{1}^j}{\sqrt{C_2^j + b_2m_2(\omega_{1}^j)^2}} \cos\left(\omega_{1}^j  t + \phi_{1}^j  -  \arctan\left[\frac{b_2\omega_{1}^j }{C_2^j}\right]\right) $$
	}}\label{eqn:structural_equations_2}
	\end{align}
	\caption{Equations giving the structural equations for the DSCM describing the mass-spring system of Example \ref{example:dscm}}
\end{figure*}

\vspace{-0.15cm}
\subsection{SOLUTIONS OF A DSCM}\label{section:solutions-to-dscm}
\vspace{-0.15cm}

Theorem \ref{theorem:structural-equations-well-defined} states that we can construct a DSCM by the described procedure. We constructed each equation by intervening on $D-1$ variables at a time. The result of this section states that the DSCM can be used to correctly model interventions on \emph{arbitrary} subsets of variables. We say that $\Eta_\mathcal{I} \in \intset_{\mathcal{I}}$ is a \emph{solution} of $\mathcal{M}$ if $\eta_i = F_i(\Eta_{\pa(i)}) \: \forall i \in \mathcal{I}$. 

\begin{theorem}\label{theorem:same-solutions}
Suppose that $(\mathcal{D},\intset)$ is structurally dynamically stable. Let $I \subseteq \mathcal{I}$, and let $\Zeta_I \in \intset_I$. Then $\mathcal{D}_{do(\mathbf{X}_I = \Zeta_I)}$ is dynamically stable if and only if the intervened SCM $\mathcal{M}_{(\mathcal{D}_{\doop(\mathbf{X}_I = \Zeta_I)})}$ has a unique solution. If there is a unique solution, it coincides with the element of $\intset_\mathcal{I}$ describing the asymptotic dynamics of $\mathcal{D}_{\doop(\mathbf{X}_I = \Zeta_I)}$.
%If $\mathcal{M}_{\mathcal{D}_{\doop(\mathbf{X}_I = \Zeta_I)}}$ does not have a unique solution, it has no solution.
\end{theorem}
%For the proof, see Section \ref{supp:theorem2proof} in the Supplementary Material.

\begin{remark}
We could also take $I = \emptyset$, in which case the above theorem applies to just $\mathcal{D}$.
\end{remark}

\vspace{-0.15cm}
\subsection{CAUSAL REASONING IS PRESERVED}\label{section:causal-reasoning-preserved}
\vspace{-0.15cm}
We have defined ways to model interventions in both ODEs and DSCMs. The following theorem and its immediate corollary proves that these notions of intervention coincide, and hence that DSCMs provide a representation to reason about the asymptotic behaviour of the ODE under interventions in $\intset$. A consequence of these results is that the diagram in Figure \ref{fig:commuting_diagram} commutes.

\begin{theorem}\label{theorem:commuting-diagram}
Suppose that $(\mathcal{D},\intset)$ is structurally dynamically stable. Let $I \subseteq \mathcal{I}$ and let $\Zeta_I \in \intset_{I}$. Then $\mathcal{M}_{(\mathcal{D}_{\mathtt{do}(\mathbf{X}_I = \pmb{\zeta}_I)})} = (\mathcal{M}_{\mathcal{D}})_{\mathtt{do}(\mathbf{X}_I = \pmb{\zeta}_I)}$.
\end{theorem}

\begin{corollary}\label{corr:double-commuting-diagram}
Suppose additionally that $J \subseteq \mathcal{I}\setminus I$ and let ${\Zeta_J \in \intset_{J}}$. Then 
\[{\left(\mathcal{M}_{(\mathcal{D}_{\mathtt{do}(\mathbf{X}_I = \pmb{\zeta}_I)})}\right)_{\mathtt{do}(\mathbf{X}_J = \pmb{\zeta}_J)} = (\mathcal{M}_{\mathcal{D}})_{\mathtt{do}(\mathbf{X}_I = \pmb{\zeta}_I, \mathbf{X}_J = \pmb{\zeta}_J)}}\,.\]
\end{corollary}
%For the proof, see Section \ref{supp:theorem3proof} in the Supplementary Material.

To summarise, Theorems \ref{theorem:structural-equations-well-defined}--\ref{theorem:commuting-diagram} and Corollary \ref{corr:double-commuting-diagram} collectively state that if $(\mathcal{D},\intset)$ is dynamically structurally stable then it is possible to derive a DSCM that allows us to reason about the asymptotic dynamics of the ODE under any possible intervention in $\intset$. 

\vspace{-0.15cm}
\subsection{RELATION TO ODEs AND DYNAMIC BAYESIAN NETWORKS}
\vspace{-0.15cm}
An ODE is capable of modelling arbitrary interventions on the system it describes. 
At the cost of only modelling a restricted set of interventions, a DSCM can be derived which describes the asymptotic behaviour of the system under these interventions. This may be desirable in cases for which transient behaviour is not important.

We now compare DSCMs to Dynamic Bayesian Networks (DBNs), an existing popular method for causal modelling of dynamical systems \citep{PGM2009}. DBNs are essentially Markov chains, and thus are appropriate for discrete-time systems. When the discrete-time Markov assumption holds, DBNs are a powerful tool capable of modelling arbitrary interventions. 
However, approximations must be made whenever these assumptions do not hold. In particular, a continuous system must be approximately discretised in order to be modelled by a DBN \citep{SokolHansen2014}.
 
By using the Euler method for numerically solving ODEs, we can make such an approximation to derive a DBN describing the system in Example~\ref{example:3-mass-spring}, leading to the discrete time equation given in \eref{eq:DBN} the Supplementary Material.
%\begin{align*}
%DBN: \left\lbrace
%\begin{array}{lll}
%X_1^{(t+1)\Delta} =  X_1(t\Delta) + \Delta \dot{X_1}(t\Delta) \\
%\dot{X_1}^{(t+1)\Delta} =  \dot{X_1}(t\Delta) + \frac{\Delta}{m_1}\Big[k_1X_2(t\Delta)  \\
%\quad\qquad\qquad- b_1 \dot{X_1}(t\Delta) - (k_0 + k_1) X_1(t\Delta) \\
%\quad\qquad\qquad  + k_0l_0 - k_1l_1 \Big]\\
%X_2^{(t+1)\Delta} =  X_2(t\Delta) + \Delta \dot{X_2}(t\Delta) \\
%\dot{X_2}^{(t+1)\Delta} =  \dot{X_2}(t\Delta) + \frac{\Delta}{m_2}\Big[k_1X_1(t\Delta)  \\
%\quad\qquad\qquad- b_2 \dot{X_2}(t\Delta) - (k_1 + k_2) X_2(t\Delta) \\
%\quad\qquad\qquad  + k_1l_1 - k_2l_2 + k_2L \Big]\\
%\\
%X_i^{(k)}(0) = (\mathbf{X}^{(k)}_0)_i  \quad k \in \{0,1\}, \: i \in \{1,2\}\,.\\
%\end{array}
%\right.
%\end{align*}
For DBNs, the main choice to be made is how fine the temporal discretisation should be.
The smaller the value of $\Delta$, the better the discrete approximation will be. 
Even if there is a natural time-scale on which measurements can be made, choosing a finer discretisation than this will provide a better approximation to the behaviour of the true system. 
The choice of $\Delta$ should reflect the natural timescales of the interventions to be considered too; for example, it is not clear how one would model the intervention $\doop\left(X_1(t) = \cos\left(\frac{2\pi t}{\Delta}\right)\right)$ with a discretisation length $\Delta$. Another notable disadvantage of DBNs is that the computational cost of
learning and inference increases for smaller $\Delta$, where computational cost becomes
infinitely large in the limit $\Delta \to 0$.

In contrast, the starting point for DSCMs is to fix a convenient set of interventions we are interested in modelling. If a DSCM containing these interventions exists, it will model the asymptotic behaviour of the system under each of these interventions \emph{exactly}, rather than approximately modelling the transient and asymptotic behaviour as in the case of a DBN. Computational cost does not relate inversely to accuracy as for DBNs, but depends on the chosen representation of the set of admitted interventions.

\vspace{-0.15cm}
\section{DISCUSSION AND FUTURE WORK}\label{section:discussion}
\vspace{-0.15cm}

The main contribution of this paper is to show that the SCM framework can be applied to reason about time-dependent interventions on an ODE in a dynamic setting. In particular, we showed that if an ODE is sufficiently well-behaved under a set of interventions, a DSCM can be derived that captures how the asymptotic dynamics change under these interventions. This is in contrast to previous approaches to connecting the language of ODEs with the SCM framework, which used SCMs to describe the stable (constant-in-time) equilibria of the ODE and how they change under intervention.

%We have extended the structural causal framework\Bernhard{we should never say that we have extended the SCM framework, since really our DSCMS are SCMs. This can backfire. Our contribution is to link ODEs and SCMs in a (simple) dynamic setting (which I consider a major contribution, don't get me wrong). I propose you rewrite this paragraph, focus on that, and shorten it a bit as well} to encompass deterministic dynamical systems exhibiting non-constant asymptotic dynamics under non-constant interventions, subject to the condition that the ODE and interventions $(\mathcal{D},\intset)$ are interventionally closed. The derived DSCMs capture the asymptotic dynamic behaviour of the dynamical systems, and allow one to predict the asymptotic dynamics resulting from interventions derived from paths in the set $\intset$.

We identify three possible directions in which to extend this work in the future. The first is to properly understand how learning DSCMs from data could be performed. This is important if DSCMs are to be used in practical applications. Challenges to be addressed include finding practical parameterizations of DSCMs, the presence of measurement noise in the data and the fact that time-series data are usually sampled at a finite number of points in time.
% \Paul{what do you think?}\Stephan{Very good! Although maybe we can use a different word for ``single component''? Or is it clear what is meant by that? To me it is not completely clear at least.} \Paul{What I'm trying to say is that the DSCM could be one part of a more complex model. I.e. if we have data and we are trying to fit a DSCM to it, we could assume that the underlying process that generated the model was deterministic, but that our data were corrupted by noise and only sampled at particular points in time. } \Joris{I think this will be lost on the reader. Let's keep it short here. I'd rather propose writing something like ``Challenges to be addressed are finding practical parameterizations of DSCMs, the presence of measurement noise in the data and the fact that time-series data are usually sampled at a finite number of points in time.''}
%
The second is to relax the assumption that the asymptotic dynamics are \emph{independent of initial conditions}, as was done recently for the static equilibrium scenario by \citet{BlomMooij_1805.06539}.
%. This rules out, for example, simple models of neural dynamics such as the FitzHugh-Namugo model which exhibits a limit cycle in the observational system \citep{fitzhugh1966mathematical}. 
%Indeed, if one were to start two systems at two different points along the limit cycle, they would remain forever out of phase, violating independence of initial condition. Intuitively, systems exhibiting limit cycles might still have dynamics that are sufficiently simple to characterise in an SCM-like framework.
%
The third extension is to move away from deterministic systems and consider Random Differential Equations \citep{BongersMooij_1803.08784}, thereby allowing to take into account model uncertainty, but also to include systems that may be inherently stochastic.

%Limitations of this approach: the independence of initial conditions assumption is severe. I think this means that the systems have to either decay to some stable equilibrium or that they have some kind of time dependent forcing within them that is independent of the state of the system. It for sure means that systems which are only dependent on the values of the variables and that have a limit cycle (such as the FitzHugh-Namugo model) cannot be used in this approach. Indeed, if you were to start at two points on the limit cycle, you would never converge to the same solution. Models exhibiting limit cycles seem like they should be simply characterisable enough to do something like build SCMs.

%\subsection{Future work}
%
%Relaxing independence of initial condition approach.

% uncomment for camera-ready copy
\vspace{-0.2cm}
\subsubsection*{ACKNOWLEDGEMENTS}
\vspace{-0.2cm}
Stephan Bongers was supported by NWO, the Netherlands Organization for Scientific Research (VIDI grant 639.072.410).
This project has received funding from the European Research Council (ERC) under the European Union's Horizon 2020 research and innovation programme (grant agreement n$^{\mathrm{o}}$ 639466).

\newpage

%\bibliography{references}

\newpage
\onecolumn
\section*{\centerline{SUPPLEMENTARY MATERIAL}}
\renewcommand{\thesection}{\Alph{section}}
\setcounter{section}{0}

\section{PROOFS}\label{supp:proofs}
\subsection{PROOF OF THEOREM 1}\label{supp:theorem1proof}
\begin{proof}
We need to show that if $\Zeta_{I_i}$ and $\Zeta'_{I_i}$ are such that $(\Zeta_{I_i})_{\pa(i)} = (\Zeta'_{I_i})_{\pa(i)} = \Eta_{\pa(i)}$, then $\eta_i = \eta'_i$. To see that this is the case, observe that the system of equations for $\mathcal{D}_{\doop(\mathbf{X}_{I_i} =  \Zeta_{I_i})}$  is given by:
\begin{align*}
\mathcal{D}_{\doop(\mathbf{X}_{I_i} =  \Zeta_{I_i})}: \left\lbrace
\begin{array}{ll}
X_j(t) = \zeta_j(t) & j \in  \mathcal{I}\setminus (\pa(i) \cup \{i\}) \,, \\
X_j(t) = \eta_j(t) & j \in  \pa(i) \,, \\
f_i(X_i,\mathbf{X}_{\mathtt{pa}(i)})(t) = 0 \quad & X_i^{(k)}(0) = (\mathbf{X}_0^{(k)})_i, \: 0\leq k \leq n_i - 1 \,. \\
\end{array}
\right.
\end{align*}
The equations for $\mathcal{D}_{\doop(\mathbf{X}_{I_i} =  \Zeta'_{I_i})}$  are similar, except with $X_j(t) = \zeta'_j(t)$  for $j \in  \mathcal{I}\setminus (\pa(i) \cup \{i\})$. % and possibly different initial conditions for $X_i$.
In both cases, the equations for all variables except $X_i$ are solved already. The equation for $X_i$ in both cases reduces to the same quantity by substituting in the values of the parents, namely
\[
f_i(X_i,\Eta_{\pa(i)})(t) = 0 \,.
\]
The solution to this equation in $\intset_i$ must be unique and independent of initial conditions, else the dynamic stability of the intervened systems $\mathcal{D}_{\doop(\mathbf{X}_{I_i} =  \Zeta_{I_i})}$ and $\mathcal{D}_{\doop(\mathbf{X}_{I_i} =  \Zeta'_{I_i})}$ would not hold, contradicting the dynamic structural stability of $(\mathcal{D},\intset)$. It follows that $\eta_i = \eta'_i$.
\end{proof}

\subsection{PROOF OF THEOREM 2}\label{supp:theorem2proof}
\begin{proof} 

By construction of the SCM, $\Eta \in \intset_\mathcal{I}$ is a solution of $\mathcal{M}_{(\mathcal{D}_{\doop(\mathbf{X}_I = \Zeta_I)})}$ if and only if the following two conditions hold:
\begin{compactitem}
\item for $i \in \mathcal{I}\setminus I$, $X_i(t) = \eta_i(t) \; \forall t$ is a solution to the differential equation $f_i(X_i, \Eta_{\pa(i)})(t) = 0$;
\item for $i \in I$, $\eta_i(t) = \zeta_i(t)$ for all $t$.
\end{compactitem}
which is true if and only if $\mathbf{X} = \Eta$ is a solution to $\mathcal{D}_{\doop(\mathbf{X}_I = \Zeta_I)}$ in $\intset_\mathcal{I}$. 
Thus, by definition of dynamic stability, $\mathcal{D}_{\doop(\mathbf{X}_I = \Zeta_I)}$ is dynamically stable with asymptotic dynamics describable by $\Eta \in \intset$ if and only if $\mathbf{X} = \Eta$ uniquely solves $\mathcal{M}_{(\mathcal{D}_{\doop(\mathbf{X}_I = \Zeta_I)})}$.
%Since $(\mathcal{D},\intset)$ is interventionally closed, $\mathcal{D}_{\doop(\mathbf{X}_I = \Zeta_I)}$ is dynamically stable with reference to $\intset_\mathcal{I}$ and thus there is exactly one element of $\intset$ describing its asymptotic dynamics. Thus there is a unique $\Eta \in \intset_\mathcal{I}$ that solves $\mathcal{M}_{(\mathcal{D}_{\doop(\mathbf{X}_I = \Zeta_I)})}$.
\end{proof}

\subsection{PROOF OF THEOREM 3}\label{supp:theorem3proof}
\begin{proof}
We need to show that the structural equations of $\mathcal{M}_{(\mathcal{D}_{\doop(\mathbf{X}_I = \pmb{\zeta}_I)})}$ and $(\mathcal{M}_{\mathcal{D}})_{\mathtt{do}(\mathbf{X}_I = \pmb{\zeta}_I)}$ are equal. Observe that the equations for $\mathcal{D}_{\doop(\mathbf{X}_I = \pmb{\zeta}_I)}$ are given by:
\begin{align*}
\mathcal{D}_{\mathtt{do}(\mathbf{X}_I = \pmb{\zeta}_I)} :\left\lbrace
\begin{array}{ll}
X_i = \zeta_i, \quad & i \in I  \,,\\
f_i(X_i, \mathbf{X}_{\pa(i)}) = 0, X_i^{(k)}(0) = (\mathbf{X}_0^{(k)})_i, \: 0\leq k \leq n_i - 1, \quad & i \in \mathcal{I} \setminus I  \,.
%&  i \in \mathcal{I} \setminus I\,. \\
\end{array}
\right.
\end{align*}
Therefore, when we perform the procedure to derive the structural equations for $\mathcal{D}_{\doop(\mathbf{X}_I = \pmb{\zeta}_I)}$, we see that:
\begin{compactitem}
\item if $i \in I$, the $i$th structural equation will simply be $X_i = \zeta_i$ since intervening on $I_i$ does not affect variable $X_i$.
\item if $i \in \mathcal{I}\setminus I$, the $i$th structural equation will be the same as for $\mathcal{M}_\mathcal{D}$, since the dependence of $X_i$ on the other variables is unchanged.
\end{compactitem}
Hence the structural equations for $\mathcal{M}_{(\mathcal{D}_{\doop(\mathbf{X}_I = \pmb{\zeta}_I)})}$ are given by:
\begin{align*}
\mathcal{M}_{(\mathcal{D}_{\doop(\mathbf{X}_I = \pmb{\zeta}_I)})} :\left\lbrace
\begin{array}{ll}
X_i = \zeta_i, \quad & i \in I  \,,\\
X_i = F_i(\mathbf{X}_{\pa(i)}), \quad & i \in \mathcal{I} \setminus I \,. \\
\end{array}
\right.
\end{align*}
and therefore $\mathcal{M}_{(\mathcal{D}_{\doop(\mathbf{X}_I = \pmb{\zeta}_I)})} = (\mathcal{M}_{\mathcal{D}})_{\mathtt{do}(\mathbf{X}_I = \pmb{\zeta}_I)}$ .
\end{proof}

\subsection{PROOF OF COROLLARY 1}\label{supp:cor1proof}
\begin{proof}
Corollary 1 follows very simply from the observation that if $(\mathcal{D},\intset)$ is structurally dynamically stable then so is $(\mathcal{D}_{\doop(\mathbf{X}_I = \pmb{\zeta}_I)}, \intset_{\mathcal{I}\setminus I} )$. The result then follows by application of Theorem 3.
\end{proof}

%\newpage
\section{DERIVING THE DSCM FOR THE MASS-SPRING SYSTEM}

Consider the mass-spring system of Example \ref{example:3-mass-spring}, but with $D\geq 1$ an arbitrary integer. We repeat the setup: 

We have $D$ masses attached together on springs. The location of the $i$th mass at time $t$ is $X_i(t)$, and its mass is $m_i$. For notational ease, we denote by $X_0=0$ and $X_{D+1} = L$ the locations of where the ends of the springs attached to the edge masses meet the walls to which they are affixed. $X_0$ and $X_{D+1}$ are constant. The natural length and spring constant of the spring connecting masses $i$ and $i+1$ are $l_i$ and $k_i$ respectively. The $i$th mass undergoes linear damping with coefficient $b_i$, where $b_i$ is small to ensure that the system is underdamped. The equation of motion for the $i$th mass ($1\leq i \leq D$) is given by:
\begin{align*}
m_i\ddot{X}_i(t) =  k_i[X_{i+1}(t) - X_i(t) - l_i]   - k_{i-1}[X_i(t) - X_{i-1}(t) - l_{i-1}] - b_i \dot{X}_i(t)
\end{align*}
so, defining 
\[
f_i(X_i,X_{i-1},X_{i+1})(t) = m_i\ddot{X}_i(t) -  k_i[X_{i+1}(t) - X_i(t) - l_i]   + k_{i-1}[X_i(t) - X_{i-1}(t) - l_{i-1}] + b_i \dot{X}_i(t)
\]
we can write the system of equations $\mathcal{D}$ for our mass-spring system as
\begin{align*}
\mathcal{D} :\left\lbrace
\begin{array}{ll}
f_i(X_i,X_{i-1},X_{i+1})(t) = 0 \quad & i \in \mathcal{I} \,. \\
\end{array}
\right.
\end{align*}
In the rest of this section we will explicitly calculate the structural equations for the DSCM derived from $\mathcal{D}$ with two different sets of interventions. First, we will derive the structural equations for the case that $\intset$ consists of all constant trajectories, corresponding to constant interventions that fix variables to constant values for all time. This illustrates the correspondence between the theory in this paper and that of \cite{MooJanSch13}. Next, we will derive the structural equations for the case that $\intset$ consists of interventions corresponding to sums of periodic forcing terms. 

\begin{subsection}{MASS-SPRING WITH CONSTANT INTERVENTIONS}\label{supp:mass-spring-constant-scm}

In order to derive the structural equations we only need to consider, for each variable, the influence of its parents on it. (Formally, this is because of Theorem \ref{theorem:structural-equations-well-defined}). Consider variable $i$. If we intervene to fix its parents to have locations $X_{i-1}(t) = \eta_{i-1}$ and $X_{i+1}(t) = \eta_{i+1}$ for all $t$, then the equation of motion for variable $i$ is given by
\begin{align*}
m_i\ddot{X}_i(t) + b_i \dot{X}_i(t) + (k_i+k_{i-1})X_{i}(t)=  k_i[\eta_{i+1} - l_i]   + k_{i-1}[ \eta_{i-1} + l_{i-1}] \,.
\end{align*}
There may be some complicated transient dynamics that depend on the initial conditions $X_i(0)$ and $\dot{X}_i(0)$ but provided that $b_i > 0$, we know that the $X_i(t)$ will converge to a constant and therefore the asymptotic solution to this equation can be found by setting $\ddot{X}_i$ and $\dot{X}_i$ to zero. Note that in general, we could explicitly find the solution to this differential equation (and indeed, in the next example we will) but for now there is a shortcut to deriving the structural equations.\footnote{This is analogous to the approach taken in \cite{MooJanSch13} in which the authors first define the Labelled Equilibrium Equations and from these derive the SCM.} The asymptotic solution is:
\begin{align*}
X_{i} = \frac{k_i[\eta_{i+1} - l_i]   + k_{i-1}[ \eta_{i-1} + l_{i-1}] }{k_i+k_{i-1}}.
\end{align*}
Therefore the $i$th structural equation is:
\begin{align*}
F_i(X_{i-1},X_{i+1}) = \frac{k_i[X_{i+1} - l_i]   + k_{i-1}[ X_{i-1} + l_{i-1}] }{k_i+k_{i-1}}.
\end{align*}
Hence the SCM for $(\mathcal{D},\intset_c)$ is:
\begin{align*}
\mathcal{M}_{\mathcal{D}} :\left\lbrace
\begin{array}{ll}
X_i = \displaystyle\frac{k_i[X_{i+1} - l_i]   + k_{i-1}[ X_{i-1} + l_{i-1}] }{k_i+k_{i-1}} \quad & i \in \mathcal{I} \,. \\
\end{array}
\right.
\end{align*}
We can thus use this model to reason about the effect of constant interventions on the asymptotic equilibrium states of the system.
\end{subsection}

\begin{subsection}{SUMS OF PERIODIC INTERVENTIONS}\label{supp:mass-spring-periodic-scm}

Suppose now we want to be able to make interventions of the form: 
\begin{equation}\label{eqn:periodic-intervention}
\doop\big( X_i(t) = A \cos(\omega t + \phi) \big) \,. %\tag{$\dagger$}
\end{equation}
Such interventions cannot be described by the DSCM derived in Section \ref{supp:mass-spring-constant-scm}. In this section we will explicitly derive a DSCM capable of reasoning about the effects of such interventions. It will also illustrate why we need dynamic structural stability.

By Theorem \ref{theorem:structural-equations-well-defined}, to derive the structural equation for each variable we only need to consider the effect on the child of intervening on the parents according to interventions of the form \eref{eqn:periodic-intervention}. Consider the following linear differential equation:
\begin{align}\label{equation:forced-de}
m\ddot{X}(t) + b \dot{X}(t) + kX(t)=  g(t)\,.
\end{align}
In general, the solution to this equation will consist of two parts---the \emph{homogeneous} solution and the \emph{particular} solution. The homogeneous solution is one of a family of solutions to the equation
\begin{align}
m\ddot{X}(t) + b \dot{X}(t) + kX(t)=  0 
\end{align}
and this family of solutions is parametrised by the initial conditions. If $b>0$ then all of the homogeneous solutions decay to zero as $t\longrightarrow\infty$. The particular solution is any solution to the original equation with arbitrary initial conditions. The particular solution captures the asymptotic dynamics due to the forcing term $g$. Equation \ref{equation:forced-de} is a linear differential equation. This means that if $X=X_1$ is a particular solution for $g = g_1$ and $X=X_2$ is a particular solution for $g = g_2$, then $X=X_1+X_2$ is a particular solution for $g=g_1+g_2$.

  In order to derive the structural equations, the final ingredient we need is an explicit representation for a particular solution to (\ref{equation:forced-de}) in the case that $g(t) = A\cos(\omega t + \phi)$. We state the solution for the case that the system is underdamped---this is a standard result and can be verified by checking that the following satisfies \eref{equation:forced-de}:
\[
X(t) = A'\cos(\omega t + \phi')
\]
where
\begin{align}\label{eqn:motion-params-transform}
 A' = \frac{A}{\sqrt{[k - m\omega^2]^2 + bm\omega^2}}\,,  && \phi' = \phi - \arctan\left[\frac{b\omega}{k - m\omega^2}\right]\,.
\end{align}
Therefore if we go back to our original equation of motion for variable $X_i$
\begin{align*}
m_i\ddot{X}_i(t) + b_i \dot{X}_i(t) + (k_i+k_{i-1})X_{i}(t)=  k_i[X_{i+1}(t) - l_i]   + k_{i-1}[ X_{i-1}(t) + l_{i-1}] 
\end{align*}
and perform the intervention 
\[
\doop(X_{i-1}(t) = A_{i-1} \cos(\omega_{i-1} t + \phi_{i-1}), X_{i+1}(t) = A_{i+1} \cos(\omega_{i+1} t + \phi_{i+1}))
\]
we see that we can write the RHS of the above equation as the sum of the three terms
\begin{align*}
g_1(t) &= k_{i-1}l_{i-1} - k_{i}l_i  \,,\\
g_2(t) &= k_{i-1}A_{i-1} \cos(\omega_{i-1} t + \phi_{i-1}) \,, \\
g_3(t) &= k_{i}A_{i+1} \cos(\omega_{i+1} t + \phi_{i+1})\,.
\end{align*}
Using the fact that linear differential equation have superposable solutions and \eref{eqn:motion-params-transform}, we can write down the resulting asymptotic dynamics of $X_i$:
\begin{align*}
X_i(t&) = \frac{k_{i-1}l_{i-1} - k_{i}l_i}{k_i + k_{i-1}}  \\
 + &\frac{k_{i-1}A_{i-1}}{\sqrt{[k_i + k_{i-1} - m_i\omega_{i-1}^2]^2 + b_im_i\omega_{i-1}^2}} \cos\left(\omega_{i-1} t + \phi_{i-1} -  \arctan\left[\frac{b_i\omega_{i-1}}{k_i + k_{i-1} - m_i\omega_{i-1}^2}\right]\right) \\
 + &\frac{k_{i}A_{i+1}}{\sqrt{[k_i + k_{i-1} - m_i\omega_{i+1}^2]^2 + b_im_i\omega_{i+1}^2}} \cos\left(\omega_{i+1} t + \phi_{i+1} -  \arctan\left[\frac{b_i\omega_{i+1}}{k_i + k_{i-1} - m_i\omega_{i+1}^2}\right]\right)\,.
\end{align*}
However, note that if we were using $\intset$ consisting of interventions of the form of equation \eref{eqn:periodic-intervention}, then we have just shown that the mass-spring system would not be structurally dynamically stable with respect to this $\intset$, since we need two periodic terms and a constant term to describe the motion of a child under legal interventions of the parents. 

This illustrates the fact that we may sometimes be only interested in a particular set of interventions that may not itself satisfy structural dynamic stability, and that in this case we must consider a larger set of interventions that \emph{does}. In this case, we can consider the modular set of trajectories generated by trajectories of the following form for each variable:
\begin{align*}
X_i(t) = \sum_{j=1}^\infty A_i^j \cos(\omega_i^j t + \phi_i^j)
\end{align*}
where for each $i$ it holds that $\sum_{j=1}^\infty |A_i^j| < \infty$ (so that the series is absolutely convergent and thus does not depend on the ordering of the terms in the sum). Call this set $\intset_{qp}$ (``quasi-periodic''). By equation \eref{eqn:motion-params-transform}, we can write down the structural equations
\begin{align*}
F_i & \left(\sum_{j=1}^\infty A_{i-1}^j \cos(\omega_{i-1}^j t + \phi_{i-1}^j), \sum_{j=1}^\infty A_{i+1}^j \cos(\omega_{i+1}^j t + \phi_{i+1}^j) \right)  \\
=&\frac{k_{i-1}l_{i-1} - k_{i}l_i}{k_i + k_{i-1}}  \\
&+ \sum_{j=1}^\infty  \frac{k_{i-1}A_{i-1}^j}{\sqrt{[k_i + k_{i-1} - m_i(\omega_{i-1}^j)^2]^2 + b_im_i(\omega_{i-1}^j)^2}} \cos\left(\omega_{i-1}^j  t + \phi_{i-1}^j  -  \arctan\left[\frac{b_i\omega_{i-1}^j }{k_i + k_{i-1} - m_i(\omega_{i-1}^j)^2}\right]\right) \\
&+\sum_{j=1}^\infty  \frac{k_{i}A_{i+1}^j}{\sqrt{[k_i + k_{i+1} - m_i(\omega_{i+1}^j)^2]^2 + b_im_i(\omega_{i+1}^j)^2}} \cos\left(\omega_{i+1}^j  t + \phi_{i+1}^j  -  \arctan\left[\frac{b_i\omega_{i+1}^j }{k_i + k_{i+1} - m_i(\omega_{i+1}^j)^2}\right]\right) \,.
\end{align*}
Since this is also a member of $\intset_{qp}$, the mass-spring system is dynamically structurally stable with respect to $\intset_{qp}$ and so the equations $F_i$ define the Dynamic Structural Causal Model for asymptotic dynamics.

\end{subsection}

\section{DYNAMIC BAYESIAN NETWORK REPRESENTATION}

By using Euler's method, we can obtain a (deterministic) Dynamic Bayesian Network representation
of the mass-spring system. For $D=2$, this yields
\begin{align}\label{eq:DBN}
DBN: \left\lbrace
\begin{array}{lll}
X_1^{(t+1)\Delta} =  X_1(t\Delta) + \Delta \dot{X_1}(t\Delta) \\
\dot{X_1}^{(t+1)\Delta} =  \dot{X_1}(t\Delta) + \frac{\Delta}{m_1}\Big[k_1X_2(t\Delta)  - b_1 \dot{X_1}(t\Delta) - (k_0 + k_1) X_1(t\Delta) + k_0l_0 - k_1l_1 \Big]\\\\
X_2^{(t+1)\Delta} =  X_2(t\Delta) + \Delta \dot{X_2}(t\Delta) \\
\dot{X_2}^{(t+1)\Delta} =  \dot{X_2}(t\Delta) + \frac{\Delta}{m_2}\Big[k_1X_1(t\Delta) - b_2 \dot{X_2}(t\Delta) - (k_1 + k_2) X_2(t\Delta)  + k_1l_1 - k_2l_2 + k_2L \Big]\\
\\
X_i^{(k)}(0) = (\mathbf{X}^{(k)}_0)_i  \quad k \in \{0,1\}, \: i \in \{1,2\}\,.\\
\end{array}
\right.
\end{align}

\end{document}